\documentclass{mindlab}
\usepackage{amsmath,amssymb}
\usepackage{wrapfig}
\usepackage{minted}
\usepackage{listings}
\usepackage{fancyvrb}
\usepackage{xcolor}
\usepackage{float}
\newfloat{listing}{htbp}{lop}
\floatname{listing}{Listing}

\newcommand{\equalcontrib}{\textsuperscript{\dag}}

\newlength\savewidth
\newcommand\shline{\noalign{\global\savewidth\arrayrulewidth
  \global\arrayrulewidth 1pt}\hline\noalign{\global\arrayrulewidth\savewidth}}
\newcolumntype{x}[1]{>{\centering\arraybackslash}p{#1pt}}
\newcolumntype{y}[1]{>{\raggedright\arraybackslash}p{#1pt}}
\newcolumntype{z}[1]{>{\raggedleft\arraybackslash}p{#1pt}}
\newcommand{\tablestyle}[2]{\setlength{\tabcolsep}{#1}\renewcommand{\arraystretch}{#2}\centering\footnotesize}
\newcommand\blfootnote[1]{%
  \begingroup
    \renewcommand\thefootnote{}\footnote{#1}%
    \addtocounter{footnote}{-1}%
  \endgroup
}

\title{Encode Once, Answer Many: \\ Internalizing Context via Parametric Flow}

\author[1]{Xinhao Zhong\equalcontrib}
\author[1]{Mengxuan Wu\equalcontrib}
\author[1]{Chonghua Liao}
\author[2]{Linxi Xie}
\author[1]{Ruicheng Feng}
\author[3]{Kangqiao Xu}
\author[1]{Yanpei Guo}
\author[1]{Hao Tan}
\author[1]{Jiaxin Hong}
\author[4]{Jason Li}
\author[5]{Panpan Zhang}
\author[5]{Ziheng Qin}
\author[1]{Qinglin Lu}
\author[6]{Yuzhang Shang}
\author[7]{Tongliang Liu}
\author[8]{Damian Borth}
\author[9]{Haggai Maron}
\author[1]{Yongxiang Hua}
\author[1]{Zhixiang Wei}
\author[1]{Jian Hu}
\author[1]{Yunfei Wu}
\author[1]{Xing Sun}
\author[1]{Xin Li}
\author[1]{Haoyu Cao}
\author[2]{Zhangyang Wang}
\author[1]{Kai Wang}

\affiliation[1]{\small{Tencent}}
\affiliation[2]{\small{UT Austin}}
\affiliation[3]{\small{JHU}}
\affiliation[4]{\small{NTU}}
\affiliation[5]{\small{NUS}}
\affiliation[6]{\small{UCF}}
\affiliation[7]{\small{University of Sydney}}
\affiliation[8]{\small{University of St. Gallen}}
\affiliation[9]{\small{Technion}}

\metadata[{\faGithub\ Github}]{
  \url{https://github.com/Cypher-Bruce/CLIK.git}
}
\metadata[{\faEnvelope\ Correspondence}]{{Haoyu Cao}, {Zhangyang Wang}, {Kai Wang}
}

\abstract{
Large language models are typically adapted to a task either by placing knowledge in the input context, which keeps it external and re-encodes it on every call, or by fine-tuning, which internalizes it but demands a costly round of optimization per task. We aim to internalize the knowledge carried by a context directly into model parameters, without per-task optimization and without keeping the context in the input at inference time. To this end, we propose \textbf{CLIK}, a context-conditioned parameter generator that maps a given context to LoRA parameters for internalizing knowledge in a single forward pass through a Transformer encoder with cross-attention, requiring no test-time gradient updates. Once the generated LoRA is inserted into a frozen base model, the model answers downstream queries from the query alone, having internalized the context into its weights rather than re-reading it. Because the adapter depends only on the context, it is generated once and reused for every query on the same document, so the cost of reading that document is amortized across queries. On document question answering, CLIK substantially improves over the no-context base model and outperforms existing hypernetwork-based generators, while remaining behind a base model that reads the context directly, and it carries over to multi-turn dialogue without any multi-turn-specific training. Analysis further shows that the context-dependent part of the generated adapter concentrates in a compact and specific substructure.
}

\begin{document}

\maketitle
\blfootnote{\dag\,Equal contribution}

\section{Introduction}
\label{sec:intro}
\begin{figure}[t]
    \centering
    \includegraphics[width=\linewidth]{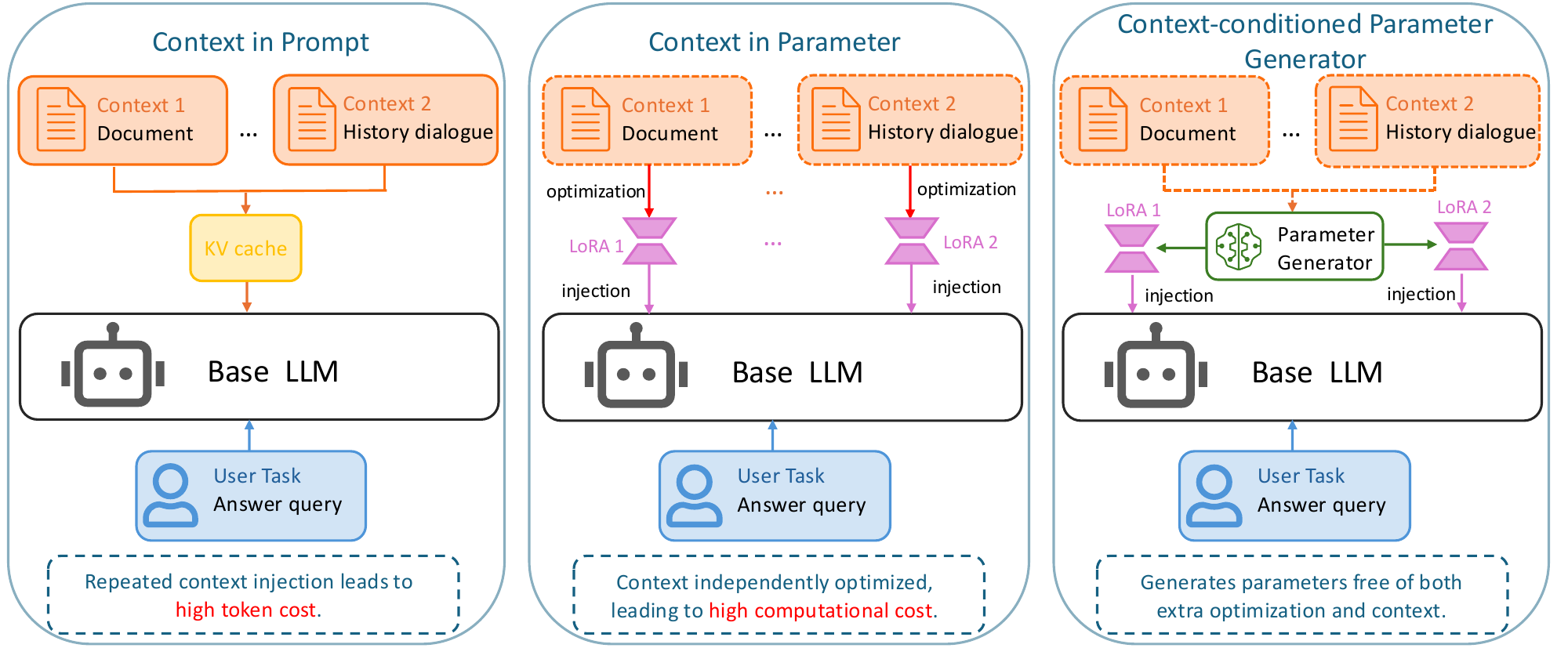}
    \caption{Three ways to supply external knowledge to a frozen LLM.
    \emph{Context in Prompt:} re-injecting the context as text incurs high token
    cost. \emph{Context in Parameter:} fine-tuning each context into a LoRA
    incurs high computational cost. \emph{Context-conditioned Parameter Generator
    (ours):} a generator maps each context to a LoRA in a single forward pass,
    free of both test-time optimization and context in the input.}
    \label{fig:introduction}
\end{figure}
Many uses of large language models (LLMs) require the model to act on a specific
body of knowledge that is not in its pretrained weights, such as a set of reference
documents, a long task instruction, or the history of an ongoing
conversation~\citep{brown2020language,laban2025llms}. The standard way to supply
this knowledge is to place it in the model's input context, as in in-context
learning and retrieval-augmented prompting, where the material is kept as text and
read by attention at generation time; this approach is flexible, requires no
training, and has become the default as context windows have grown. A second route
is to write the knowledge into the weights by fine-tuning, so that the model answers
without the context in its input~\citep{ouyang2022training,zhang2026instruction}:
parameter-efficient methods such as LoRA make each update cheap to
store~\citep{hu2022lora}, and context distillation trains the model to reproduce,
without the context, the outputs it would have produced with
it~\citep{snell2022learning,kujanpaa2024efficient}.

Keeping the knowledge in the input runs into two well-documented problems. The first
is that quality does not scale with length, because as inputs grow, models
increasingly fail to use the relevant information and their accuracy degrades, an
effect reported as lost-in-the-middle behavior, weak long in-context learning, and
context rot~\citep{liu2024lost,li2024long,hong2025context}. The second is that cost
grows with length, because self-attention is quadratic in the input and the
key--value cache that must be stored and re-read at every decoding step grows
linearly with the number of context tokens, so both latency and memory rise with the
amount of knowledge supplied~\citep{ge2024model}. Prompt- and cache-compression
methods relieve this overhead by shortening the input or shrinking the
cache~\citep{jiang2023llmlingua,pan2024llmlingua,ge2024model}, but the knowledge
still lives outside the parameters and must be re-supplied and re-read for every new
query, so it is supplied to the model but never internalized by it.

Writing the knowledge into the weights instead internalizes it directly, but each new
context needs its own optimization run, which is slow and hyperparameter-sensitive and
has to be repeated whenever the knowledge changes. This becomes impractical when
contexts are numerous, user-specific, or arrive continually~\citep{li2025beyond}.
Fine-tuning therefore internalizes knowledge but cannot adapt on the fly.

These two routes trade off in opposite directions (\Cref{fig:introduction}).
Context-based adaptation is
instant but never internalizes its knowledge, while fine-tuning internalizes the
knowledge but is too costly to update for each new context. Hypernetworks~\citep{ha2016hypernetworks}
offer a way to keep the advantages of both, since one network can emit the weights
of another and thus turn an input into an adapter in a single forward pass, with no
test-time optimization and nothing left in the context
window~\citep{chen2025generative,abdalla2025zhyper,liang2026drag}. Recent work brings
this idea to LLMs by mapping a task description or a short context to a LoRA
adapter~\citep{charakorn2026doc,liu2026shine}. Turning a long context into a faithful
adapter is harder than a naive application of this idea, because the adapter is
high-dimensional and existing generators emit its weights in independent
fragments that ignore dependencies across layers and modules, so the longer and
richer the context, the harder it is to capture in one coherent set of weights.

We propose to directly generate context-conditioned LoRA for internalizing knowledge (\textbf{CLIK}), a parameter generator built for this regime that maps a
context directly to LoRA parameters in a single forward pass. CLIK encodes the
context with the frozen backbone itself and uses a Transformer that attends from the
context to the adapter through cross-attention, generating the full set of LoRA
weights jointly rather than fragment by fragment, so that dependencies across layers
and modules are modeled as the knowledge is internalized. The generated adapter is
inserted into the frozen model, which then answers from the query alone, without
further access to the original context.

This design is particularly suited to multi-turn interaction, the setting where
existing paradigms struggle most~\citep{laban2025llms,li2025beyond}. Once a context is
internalized into the adapter, it no longer has to occupy the input on later turns,
and because the adapter depends only on the context, it is generated once and reused
for every subsequent query on the same document, so the cost of reading that document
is amortized across queries rather than paid again at each turn.
CLIK therefore combines the immediacy of in-context learning with the persistence of
parameterized knowledge.

Our main contributions are summarized as follows:
\begin{itemize}
    \item \textbf{A context-to-adapter generator that internalizes knowledge jointly.}
    We propose CLIK, which maps a context to a full LoRA adapter in a single forward
    pass via cross-attention over the backbone's all-layer representations. Unlike
    generators that emit weights as independent fragments, it models dependencies across
    layers and modules, internalizing a rich context into one coherent adapter with no
    test-time optimization.
    \item \textbf{Strong empirical results.} On single-hop and multi-hop document
    question answering, CLIK substantially outperforms the no-context base model and
    prior hypernetwork-based generators, while remaining behind a base model with full
    context access, and it transfers to multi-turn dialogue without
    multi-turn-specific training.
    \item \textbf{Analysis of how context is internalized.} Through intervention and
    distributional analysis of the generated adapters, we show that the context-dependent
    part of the adapter concentrates in a compact and specific substructure of the LoRA.
\end{itemize}
\section{Related Work}
\label{sec:related}

\subsection{Context Internalization}
Context internalization transfers the information a model would otherwise read from
its context into a form it can use directly, so that downstream queries are answered
without re-processing the original context. Existing methods divide by whether they
require parameter optimization. \emph{Training-free} methods keep the weights fixed
and act on the input: in-context learning places the material in the
prompt~\citep{brown2020language}, while prompt- and KV-cache compression shorten the
input or shrink the cached activations to cut test-time
cost~\citep{jiang2023llmlingua,pan2024llmlingua,ge2024model}. In all cases the
knowledge stays outside the parameters and must be re-supplied for every query.
\emph{Optimization-based} methods instead move the information into the weights:
supervised fine-tuning absorbs a corpus but needs curated
data~\citep{zhang2026instruction,hu2022lora}, and context distillation trains the
model to reproduce, without the context, the outputs it would have produced with
it~\citep{askell2021general,snell2022learning,kujanpaa2024efficient}. These avoid
re-reading the context at inference, but each new context triggers a separate
optimization loop that is too slow for interactive use. CLIK keeps the benefits of
both by amortizing this per-context optimization into a single learned mapping.

\subsection{Hypernetworks}
A hypernetwork produces the parameters of another
network~\citep{ha2016hypernetworks}, turning adaptation into a forward pass rather
than a gradient search. For LLMs, this means adapting a frozen model without long
prompts or fine-tuning, typically by generating a low-rank LoRA
adapter~\citep{hu2022lora} from a task description or a
context~\citep{chen2025generative,abdalla2025zhyper}. The central difficulty is the
high-dimensional output: many designs emit weight fragments independently and
concatenate them, ignoring dependencies across the adapter. Closest to us,
Doc-to-LoRA~\citep{charakorn2026doc} meta-learns the context-distillation process into
a per-layer independent hypernetwork that maps a context to a LoRA in a single pass, and
SHINE~\citep{liu2026shine} instead inserts and trains an additional LoRA on the base model together with learned memory tokens appended to the context, whose hidden states at every layer form a fixed-length memory that serves as the input to a subsequent hypernetwork. CLIK follows this
in-context line but uses a Transformer that attends from the context to the adapter
through cross-attention, generating the full set of LoRA weights jointly in one
forward pass. Beyond text, the same input-to-weights mechanism spans modalities and
tasks: text-guided image editing~\citep{tencent2026hy}, large-scale parameter
generation by diffusion~\citep{wang2025recurrent,wang2024neural}, reinforcement-learned parameter
adaptation for video-anomaly detection~\citep{jacob2026copra}, internalizing a video
or a codebase into LoRA parameters~\citep{suri2026video2lora,hotsko2026code2lora}, and
agentic settings where one agent edits another's weights rather than giving textual
advice~\citep{bao2026good}. Our context-to-LoRA formulation is one instance of this
broader trend of treating weight space as a first-class generative
modality~\citep{wang2026weightspace}, specialized to LLM context adaptation.
\begin{figure}
    \centering
    \includegraphics[width=\linewidth]{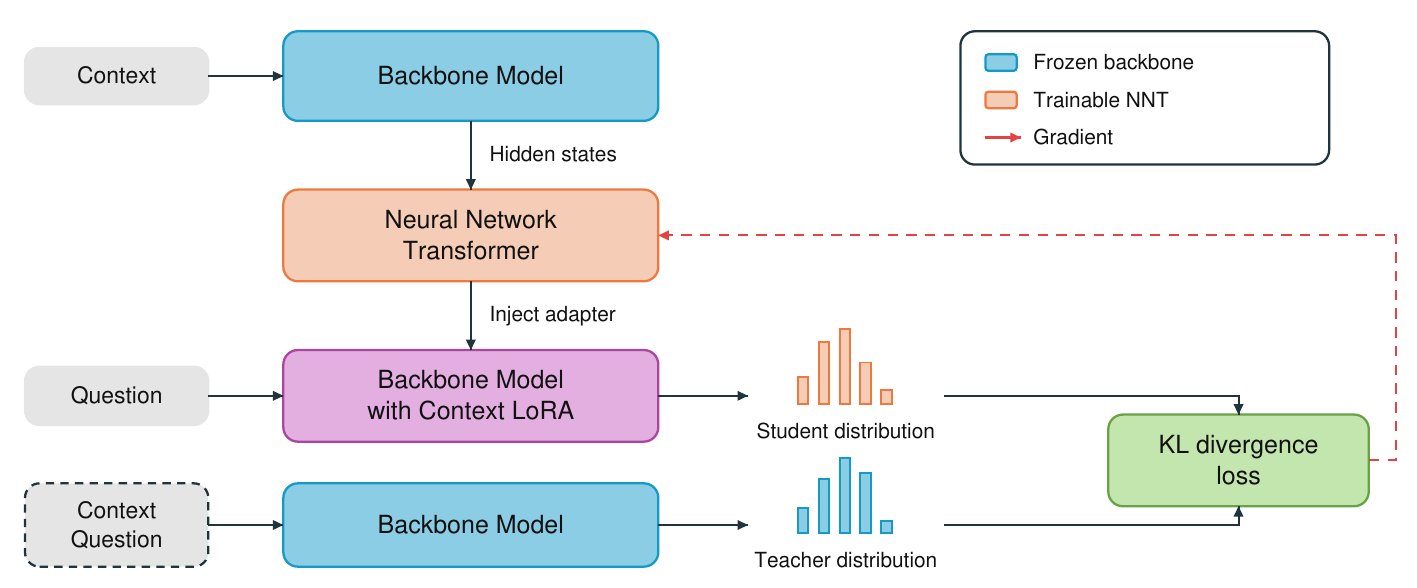}
    \caption{
        \textbf{Overview of CLIK.}
        A frozen language model encodes the context, and a learned ELMo-style weighting of its all-layer hidden states conditions a trainable generator that produces a context-specific LoRA adapter in a single forward pass. The adapter is injected into the same frozen backbone, which then answers the question from a context-free prompt (the \emph{student}). We train the parameter generator by distilling an \emph{in-context teacher}, the identical frozen model that reads context and question together, into this student. The objective is a temperature-scaled KL divergence between the teacher and student next-token distributions over the shared answer tokens. The backbone is frozen and shared across all paths, and only the generator is trainable. Gradients (dashed red) flow back through the injected adapter to the generator alone.
    }
    \label{fig:overview}
\end{figure}

\section{Method}
We present \textbf{CLIK}, a framework that internalizes a context passage into the parameters of a frozen language model by generating a low-rank adapter from that context in a single forward pass. We study knowledge-grounded question answering, where each example consists of a context passage, a question, and a target answer. Placing the context in the prompt re-encodes it for every query and inflates the input, so we instead internalize the context into the weights of the frozen language model and produce the answer from a context-free prompt that contains only the question. We learn a generator network that maps a context to a low-rank adapter (LoRA)~\citep{hu2022lora} for the model's feed-forward projections. The language model stays frozen, and the parameter generator is the only trainable component.

We design the generator around three choices. First, to obtain a strong representation of the context without training a separate encoder, we reuse the frozen backbone itself, so the context is encoded by the same model it will adapt and the conditioning signal lives natively in the model's own representation space. Second, to capture information ranging from low-level surface form to high-level semantics, we draw on the backbone's hidden states from all of its layers and combine them through a learned weighting. Third, to produce the high-dimensional adapter efficiently, we treat the target adapter as a structured sequence of weight tokens and train a neural network transformer that performs message passing among them and predicts the adapter.

As illustrated in \Cref{fig:overview}, CLIK proceeds in three stages:
\begin{enumerate}
    \item \textbf{Context Encoding.} The frozen backbone reads the context in a single forward pass, yielding a sequence of per-token hidden states at every layer. We combine these multi-layer states into a single conditioning sequence through a learnable, ELMo~\citep{peters2018deep}-style weighted sum, retaining the full token sequence as input to the generator.
    \item \textbf{Parameter Generation.} The generator takes the context hidden states as input and predicts the target adapter. It represents the adapter as a two-dimensional grid of weight tokens, indexed by backbone layer and by within-layer position, and applies factored self-attention along each axis, which lets representations for different layers exchange information (including from deep layers back to shallow ones), together with cross-attention to the context hidden states. 
    \item \textbf{Downstream Answering.} The generated adapter is injected into the frozen backbone, which then answers the question from a context-free prompt, one that no longer contains the context.
\end{enumerate}

\subsection{Training Data Construction}
\label{sec:data}

Since the backbone must answer from the generated adapter alone, the training corpus consists of self-contained passages to be internalized. We draw context passages from FineWeb-Edu~\citep{penedo2024fineweb}, a large pretraining corpus. We prefer a pretraining-scale, general-domain corpus over a task-specific QA dataset for two reasons. Its diversity drives the generator toward a generic context-to-parameter mapping rather than a single benchmark's distribution, and because it is tied to no benchmark, the generator is never trained on the evaluation tasks themselves, and its long free-form answers keep every evaluation task out of distribution; \Cref{sec:counterfactual} measures the effect of adding in-domain training data. Our ablation confirms this choice, as FineWeb-Edu surpasses task-specific or smaller corpora such as SQuAD, MS~MARCO, and TransMLA (\Cref{tab:ablation}).

For each passage we synthesize several question--answer pairs as supervision, which teaches the generator both what to internalize and how it must be recalled while probing different facts within one context. We generate these pairs with the same frozen backbone under a fixed system prompt, so the supervision follows the model's native style, and this alignment is what makes the context-free student a faithful approximation of the in-context teacher it matches. Because our goal is to reproduce in-context behavior from a context-free prompt, we impose no constraint on answer length or style, following Doc-to-LoRA~\citep{charakorn2026doc}. For the same reason of generality, neither the teacher nor the student prompt contains an answer-format instruction: the student prompt consists of the question alone and the teacher prompt of the context followed by the question, both wrapped in the backbone's chat template.

\subsection{Context Encoding}
\label{sec:context-encoding}

The first stage turns the context into a conditioning signal for the generator. Because the frozen backbone is itself a strong encoder, we run it over the context once and extract all intermediate hidden states. A backbone with $L$ layers produces $L+1$ hidden states (the embedding output and each layer's output), denoted $\mathbf{H}_0, \dots, \mathbf{H}_L$, each of shape $n \times d$ for $n$ context tokens and hidden dimension $d$. The signal useful for adaptation is spread across depth, with shallow layers carrying surface cues and deeper ones more abstract content, so we condition on all layers rather than a single one and collapse the $L+1$ states into one representation with an ELMo-style learned weighting,
\begin{equation}
    \bar{\mathbf{H}} \;=\; \gamma \sum_{\ell=0}^{L} a_\ell\, \mathbf{H}_\ell,
    \qquad
    a_\ell \;=\; \frac{\exp(w_\ell)}{\sum_{k=0}^{L} \exp(w_k)},
    \label{eq:elmo}
\end{equation}
where the per-layer weights $w_0, \dots, w_L$ and scalar $\gamma$ are learned jointly with the generator. Initializing every $w_\ell$ to zero and $\gamma$ to one starts the combination as a uniform average, and the model then learns which depths to emphasize. The aggregated $\bar{\mathbf{H}}$ retains the full length-$n$ token sequence rather than being pooled, which gives the generator position-specific information, and a linear projection maps each position from $d$ to the generator hidden dimension $d_g$ to form the conditioning sequence. This learned mixture avoids prematurely committing to where useful information resides, and our ablation confirms its value, as the last layer, a middle layer, or a uniform mean all underperform it (\Cref{tab:ablation}).

\subsection{Parameter Generator Pipeline}
\label{sec:generator}

Our parameter generator follows the Neural Network Transformer introduced in HY-WU~\citep{tencent2026hy}, which we summarize here and adapt to our setting. The generator emits a complete LoRA adapter in a single pass. Because the adapted projections have different input and output dimensions across modules and layers, their factor matrices cannot be produced by one fixed-size head without either a large parameter count or the loss of their structure. We instead generate the adapter as a sequence of uniformly shaped weight tokens, refine these tokens with a Transformer that attends to the conditioning sequence from the previous stage, and detokenize them back into factor matrices.

\paragraph{Weight tokenization.} The rank $r$ is shared by every adapter, whereas the channel dimensions are not, so every factor matrix has $r$ times a module-specific number of entries. Therefore, we flatten each factor matrix and split it into consecutive chunks of $r \cdot c$ entries, padding the last chunk when needed, and reshape each chunk into an $r \times c$ token. Each factor matrix thus becomes a set of tokens of identical shape. Gathering the tokens of all adapted projections within a layer and stacking them over the $L$ layers represents the adapter as a four-axis array indexed by layer, within-layer position, and the two axes of each token. Detokenization inverts this layout to recover the factor matrices.

\paragraph{Factorized attention.} Full attention over the flattened array scales quadratically with its length and is impractical for an all-layer adapter, so each block attends along one axis at a time. An across-layer step connects tokens that occupy the same role at different depths, and a within-layer step connects the tokens of a single layer. Together they model how each projection varies with depth and how co-located projections interact. A cross-attention step then reads the conditioning sequence, which is the only way context information enters the tokens, since the tokens themselves are initialized from learned layer and within-layer position embeddings. We apply a separate rotary position encoding \citep{su2023roformerenhancedtransformerrotary} to each axis so that every token retains its layer and position identity, while interleaving the steps with normalization and feed-forward sublayers.

\paragraph{Injection.} After detokenization, each generated pair is inserted additively into its projection: a projection that computes $\mathbf{W}\mathbf{x}$ instead computes
\begin{equation}
    \mathbf{W}\mathbf{x} \;+\; \frac{\alpha}{r}\, \mathbf{B}\mathbf{A}\, \mathbf{x},
    \label{eq:inject}
\end{equation}
with adapter scale $\alpha$. Each context yields its own $\mathbf{A}$ and $\mathbf{B}$, so the modification is applied per input, and gradients from the downstream loss reach the generator only through these terms.

\paragraph{Training objective.} We train the generator by context distillation~\citep{hinton2015distillingknowledgeneuralnetwork}, matching a context-free student to an in-context teacher. Both are the same frozen backbone: for a triple $(c,q,\mathbf{a})$ with answer $\mathbf{a}=(a_1,\dots,a_N)$, the teacher reads the context in its prompt while the student carries it through the generated adapter $\Phi(c)$. At answer position $t$ their next-token distributions, each a softmax at temperature $\tau$, are
\begin{equation}
    p_t = p_{\mathrm{LM}}^\tau(\cdot \mid \mathbf{a}_{<t}, q, c), \qquad
    q_t = p_{\mathrm{LM}}^\tau(\cdot \mid \mathbf{a}_{<t}, q;\, \Phi(c)),
    \label{eq:ts}
\end{equation}
and since the answer tokens are shared, they align position by position. Treating the teacher (computed with gradients disabled) as a fixed target, we minimize their temperature-scaled KL divergence over the $N$ answer tokens,
\begin{equation}
    \mathcal{L} = \frac{\tau^2}{N} \sum_{t=1}^{N} D_{\mathrm{KL}}\!\left(p_t \,\|\, q_t\right),
    \label{eq:distill}
\end{equation}
where the direction $p_t \,\|\, q_t$ draws the student toward the teacher's full softened distribution rather than its top token alone, and $\tau^2$ compensates for the temperature-induced gradient scaling. Only the generator is updated, and the gradient reaches it solely through $\Phi(c)$. As a simpler alternative that needs no teacher pass, the student can instead be trained with plain next-token cross-entropy on $\mathbf{a}$, but we adopt distillation by default. An optional penalty on the generated weights' magnitude is available but kept disabled in our experiments.

\paragraph{Inference.} At inference time the context is consumed entirely by the generator and never enters the prompt. Given a context $c$, we run the frozen backbone once to obtain the aggregated conditioning sequence $\bar{\mathbf{H}}(c)$ of \Cref{eq:elmo}, pass it through the generator $G_\theta$ to produce the full set of adapter factors in a single forward pass,
\begin{equation}
    \Phi(c) \;=\; \bigl\{(\mathbf{A}^{(m)}, \mathbf{B}^{(m)})\bigr\}_{m} \;=\; G_\theta\!\bigl(\bar{\mathbf{H}}(c)\bigr),
    \label{eq:infer-gen}
\end{equation}
and inject $\Phi(c)$ into the backbone's feed-forward projections through \Cref{eq:inject}. The adapted model then answers a question $q$ from a context-free prompt by greedy decoding from the question alone,
\begin{equation}
    \hat{a}_t \;=\; \operatorname*{arg\,max}_{a \in \mathcal{V}} \; p_{\mathrm{LM}}\!\bigl(a \,\big|\, \hat{\mathbf{a}}_{<t},\, q;\, \Phi(c)\bigr), \qquad t = 1, 2, \dots,
    \label{eq:infer}
\end{equation}
until an end-of-sequence token is produced, where $\mathcal{V}$ is the vocabulary and $p_{\mathrm{LM}}(\,\cdot\,;\Phi(c))$ denotes the frozen backbone with the generated adapter injected. Because $\Phi(c)$ depends only on the context, it is produced once and reused for every question about the same context, so the cost of internalization is paid a single time and amortized across queries, and the backbone itself never re-reads the context.

\section{Experiments}
\label{sec:experiments}

\subsection{Experimental Settings}

\paragraph{Implementation Details.} We use Qwen3-4B-Instruct~\citep{yang2025qwen3}
as the frozen base model throughout our experiments. CLIK is built from a stack of
$12$ Transformer blocks and produces LoRA adapters of rank $16$ for the FFN layers
of the base model, i.e., for the gate, up, and down projections of all $36$ layers,
which amounts to $21.2$M generated parameters ($40.5$\,MiB in BF16) per context. The
generator itself has $1.01$B parameters, of which $604$M are in its attention
sublayers and $403$M in its feed-forward sublayers (\Cref{tab:pg_config}); because this is a substantial
capacity, we additionally compare against baselines at matched capacity, and report
their training compute, in \Cref{sec:controlled}. For training data, we use the FineWeb-Edu 10B
subset~\citep{penedo2024fineweb} as the context corpus, and for each context we
prompt the backbone to generate $5$ question--answer pairs, which together form the
final training set. The context fed to the generator is truncated to a maximum
sequence length of $2048$ tokens during training. We train CLIK for $2$ epochs with
the AdamW optimizer, using a peak learning rate of $5\times10^{-5}$ with linear
warmup followed by decay. All experiments are run on multiple nodes of $64$ GPUs.

\paragraph{Baselines.} We compare CLIK against the following baselines.
\textbf{Base} denotes the frozen backbone that receives only the query, without the
corresponding context. \textbf{Base + Context} additionally provides the full
context in the input and serves as a gold reference, since the model has direct
access to all relevant information at inference time. We further consider two
methods that internalize the context into parameters. \textbf{Gen
Adapter}~\citep{chen2025generative} internalizes the context into LoRA parameters at
test time, and \textbf{Doc-to-LoRA}~\citep{charakorn2026doc} is a recent approach
that trains a hypernetwork to generate, layer by layer, a separate LoRA module for
each layer of the frozen backbone from a given context in a single forward pass,
which is the setting closest to ours. The original Doc-to-LoRA generates rank-$8$
adapters for the down projections only and is trained on a mixture that also includes
SQuAD, DROP, and ROPES, so we re-train it on the same FineWeb-Edu question--answer
pairs as CLIK with its own context-distillation objective, and let it generate the
same rank-$16$ adapters on the gate, up, and down projections of all $36$ layers.

Recent work \textbf{SHINE}~\citep{liu2026shine} also trains a hypernetwork that
generates a LoRA from the backbone's own context representation and is
architecturally the closest method to ours. The original SHINE uses Qwen3-8B as the
backbone, pretraining on TransMLA data followed by instruction fine-tuning on
question-answering collections, and rank-$8$ adapters on both the attention and the
feed-forward projections, so in \Cref{tab:main} we re-train it with Qwen3-4B-Instruct
as the backbone and the FineWeb-Edu question--answer pairs as training data, with its
own next-token prediction objective, and let it generate the same rank-$16$ adapters
as CLIK and Doc-to-LoRA. Because SHINE reads each memory state directly as a slice of
the adapter, the number of memory states per layer must cover the per-layer adapter
size; as $589{,}824$ is not a multiple of the hidden size $2560$, we use $231$ memory
states per layer and discard the $1{,}536$ surplus entries.
All generators in \Cref{tab:main} thus share the backbone, the training data, and
the generated adapter, but still differ in generator size, training objective, and
training compute, so \Cref{sec:controlled} additionally scales both baselines to the
parameter count of CLIK and CLIK down to a much smaller generator.
For the cost analysis in \Cref{sec:efficiency} we further include
\textbf{prefix-cached Base + Context}~\citep{kwon2023efficient,zheng2024sglang}, KV-cache
compression with \textbf{SnapKV}~\citep{li2024snapkv}, and prompt compression with
\textbf{LLMLingua-2}~\citep{pan2024llmlingua}.

\paragraph{Evaluation Setup.} We evaluate all methods on two groups of
question-answering datasets: the single-hop datasets SQuAD~\citep{rajpurkar2016squad},
ROPES~\citep{lin2019reasoning}, and DROP~\citep{dua2019drop}, and the multi-hop
datasets HotpotQA~\citep{yang2018hotpotqa}, MuSiQue~\citep{trivedi2022musique},
and 2WikiMultihopQA~\citep{ho2020constructing}. Performance is measured with
F1 score and ROUGE-L. During inference, the context is removed from the input for
all methods except the Base + Context setting, so that every other method must
answer from the query alone, relying solely on the knowledge internalized into the
parameters.

\begin{table*}[t]
\centering
\caption{Main results on the single-hop and multi-hop question-answering benchmarks
with Qwen3-4B-Instruct. We report F1 and ROUGE-L under two evaluation protocols:
without and with an instruction that constrains the output format; the backbone's
chat template is applied in both. Except for
\textbf{Base + Context}, which serves as a gold reference with direct access to the
context, all methods answer from the query alone after internalizing the context
into parameters. The best result in each column is shown in \textbf{bold} and the
second best is \underline{underlined} (Base + Context is excluded as a gold reference).}
\label{tab:main}
\resizebox{\textwidth}{!}{%
\tablestyle{5pt}{1.2}
\begin{tabular}{l ccc ccc ccc ccc}
\shline
\noalign{\vspace{1pt}}
 & \multicolumn{6}{c}{F1 Score} & \multicolumn{6}{c}{ROUGE-L} \\
\cmidrule(lr){2-7} \cmidrule(lr){8-13}
Model & \multicolumn{3}{c}{Single-Hop QA} & \multicolumn{3}{c}{Multi-Hop QA}
 & \multicolumn{3}{c}{Single-Hop QA} & \multicolumn{3}{c}{Multi-Hop QA} \\
\cmidrule(lr){2-4} \cmidrule(lr){5-7} \cmidrule(lr){8-10} \cmidrule(lr){11-13}
 & SQuAD & DROP & ROPES & HotpotQA & MuSiQue & 2WikiMQA
      & SQuAD & DROP & ROPES & HotpotQA & MuSiQue & 2WikiMQA \\
\noalign{\vspace{1pt}}
\shline
\noalign{\vspace{1pt}}
\rowcolor{gray!10} \multicolumn{13}{c}{\textit{without Format Instruction}} \\
\noalign{\vspace{1pt}}
Base + Context & 22.6 & 5.5 & 2.4 & 10.3 & 3.6 & 7.0 & 22.6 & 5.5 & 2.4 & 10.1 & 3.6 & 7.0 \\
\hline
\noalign{\vspace{1pt}}
Base          & 2.2  & 2.6 & 1.8 & 2.3 & 1.1 & 1.2 & 2.2 & 2.6 & 1.8 & 2.3 & 1.0 & 1.2 \\
Gen Adapter   & 18.5 & 2.4 & 2.9 & 3.2 & 2.0 & 4.5 & 18.2 & 2.2 & 2.9 & 3.0 & 2.0 & 4.5 \\
Doc-to-LoRA   & \underline{22.3} & \underline{5.0} & 3.1 & 6.5 & \underline{2.8} & 6.8 & \underline{21.1} & \underline{5.0} & 3.1 & 6.5 & \underline{2.6} & 6.2 \\
SHINE & 20.5  & 4.4 & \underline{3.8} & \underline{7.0} & 2.1 & \underline{7.5} & 20.4 & 4.4 & \underline{3.7} & \underline{7.0} & 2.0 & \underline{7.3} \\
\midrule
 \textbf{CLIK} & \textbf{24.5} & \textbf{5.3} & \textbf{4.2} & \textbf{7.9} & \textbf{3.4} & \textbf{8.2} & \textbf{24.5} & \textbf{5.3} & \textbf{4.2} & \textbf{7.6} & \textbf{3.4} & \textbf{7.4} \\
\noalign{\vspace{1pt}}
\shline
\noalign{\vspace{1pt}}
\rowcolor{gray!10} \multicolumn{13}{c}{\textit{with Format Instruction}} \\
\noalign{\vspace{1pt}}
Base + Context & 71.1 & 37.1 & 57.1 & 65.2 & 24.5 & 47.5 & 70.2 & 37.1 & 57.1 & 65.2 & 23.2 & 45.5 \\
\hline
\noalign{\vspace{1pt}}
Base          & 15.1 & 7.6 & \underline{35.6} & 20.7 & \underline{6.5} & 15.5 & 14.4 & 7.2 & \underline{34.4} & 19.2 & 6.2 & 14.2 \\
Gen Adapter   & 30.1 & 5.2 & 7.8  & 16.5 & 5.6 & 9.2 & 30.1 & 5.0 & 7.6 & 15.8 & 5.4 & 8.8 \\
Doc-to-LoRA   & 29.6 & 7.8 & 12.1 & 15.3 & 6.4 & 10.2 & 28.8 & 7.8 & 11.8 & 14.2 & \underline{6.3} & 10.2 \\
SHINE   & \underline{35.2} & \underline{10.4} & 34.4 & \underline{22.2} & 6.0 & \underline{18.7} & \underline{35.2} & \underline{10.3} & 34.2 & \underline{22.1} & 6.0 & \underline{18.5} \\
\midrule
 \textbf{CLIK} & \textbf{46.3} & \textbf{14.3} & \textbf{49.4} & \textbf{28.7} & \textbf{12.1} & \textbf{29.2} & \textbf{45.1} & \textbf{14.3} & \textbf{48.8} & \textbf{28.5} & \textbf{12.1} & \textbf{28.8} \\
\noalign{\vspace{1pt}}
\shline
\end{tabular}%
}
\end{table*}

\begin{table*}[t]
\centering
\caption{Decomposition of the single-hop results on SQuAD, DROP, and ROPES into
Precision, Recall, and F1 (P / R / F1), reported under both evaluation protocols. Relative to
Doc-to-LoRA, CLIK attains substantially higher precision, at comparable recall on
SQuAD and lower recall on DROP and ROPES with the format instruction, yielding a markedly
better precision--recall balance.}
\label{tab:split}
\resizebox{0.8\textwidth}{!}{%
\tablestyle{6pt}{1.2}\normalsize
\begin{tabular}{l ccc ccc ccc}
\shline
\noalign{\vspace{1pt}}
 & \multicolumn{3}{c}{SQuAD} & \multicolumn{3}{c}{DROP} & \multicolumn{3}{c}{ROPES} \\
\cmidrule(lr){2-4} \cmidrule(lr){5-7} \cmidrule(lr){8-10}
Model & Precision & Recall & F1 Score & Precision & Recall & F1 Score & Precision & Recall & F1 Score \\
\noalign{\vspace{1pt}}
\shline
\noalign{\vspace{1pt}}
\rowcolor{gray!10} \multicolumn{10}{c}{\textit{with Format Instruction}} \\
\noalign{\vspace{1pt}}
Base + Context & 67.4 & 89.6 & 71.1 & 32.6 & 59.3 & 37.1 & 52.7 & 90.1 & 57.1 \\
\hline
\noalign{\vspace{1pt}}
Doc-to-LoRA    & 20.2 & 79.1 & 29.6 & 4.6 & \textbf{49.9} & 7.8 & 8.0 & \textbf{90.3} & 12.1 \\
 CLIK & \textbf{39.4} & \textbf{81.7} & \textbf{46.3} & \textbf{11.1} & 43.9 & \textbf{14.3} & \textbf{45.6} & 79.2 & \textbf{49.4} \\
\noalign{\vspace{1pt}}
\shline
\noalign{\vspace{1pt}}
\rowcolor{gray!10} \multicolumn{10}{c}{\textit{without Format Instruction}} \\
\noalign{\vspace{1pt}}
Base + Context & 14.2 & 93.9 & 22.6 & 3.0 & 77.7 & 5.5 & 1.2 & 98.9 & 2.4 \\
\hline
\noalign{\vspace{1pt}}
Doc-to-LoRA    & 14.2 & 82.1 & 22.3 & 3.0 & 54.0 & 5.0 & 1.6 & 96.3 & 3.1 \\
 CLIK & \textbf{15.6} & \textbf{86.6} & \textbf{24.5} & \textbf{3.1} & \textbf{57.5} & \textbf{5.3} & \textbf{2.2} & \textbf{97.2} & \textbf{4.2} \\
\noalign{\vspace{1pt}}
\shline
\end{tabular}
}
\end{table*}

\subsection{Main Results}
\label{sec:main_results}
As the reference answers in our evaluation datasets are short, the
formatting of a model's response substantially influences span-overlap metrics. We
accordingly evaluate under two protocols: one in which the question is given alone,
and one in which it is preceded by an instruction that explicitly constrains the
output format (e.g., \emph{``Answer the following question. Output only the answer
and do not output any other words.''}). In both protocols the prompt is wrapped in
the backbone's chat template. The two protocols are presented as separate blocks in
\Cref{tab:main}, and the F1 and ROUGE-L scores display consistent trends. CLIK
is trained without an answer-format instruction (\Cref{sec:data}), so the first
protocol matches its training format, whereas the second is out of distribution for
the generator.

Under the protocol without the format instruction, all methods, including Base + Context,
produce long free-form answers, so the scores are low and mostly
reflect output length and format. CLIK surpasses all context-internalization
baselines and, on several datasets, exceeds the Base + Context reference, but given
how strongly this protocol is dominated by output format, we do not take this as
evidence that CLIK matches direct access to the context. Under the protocol with the
format instruction, where the base model can fully exploit it,
CLIK is the best context-free method on all six datasets, with a clear margin in
average F1 over both SHINE and Doc-to-LoRA, and it improves substantially over
Doc-to-LoRA across all single-hop datasets. It nonetheless remains clearly behind
Base + Context on every dataset, with the smallest gap on ROPES and the largest on
HotpotQA. Part of this gap stems
from the train--test mismatch in prompt format noted above, which affects the
generator but not Base + Context. The margin over Doc-to-LoRA shows that CLIK
preserves the base model's instruction-following capability after internalization
rather than merely reproducing surface-level content, even though the
output-format instruction is never seen during training.
\Cref{sec:clikb} keeps this generator fixed and instead reduces the number of
parameters it generates for each context.

\Cref{tab:split} further characterizes answer quality through precision and
recall. The context-internalization baselines achieve high recall but low
precision, producing outputs that cover the relevant content but remain loosely
scoped. CLIK departs from this pattern and substantially improves precision, at
comparable recall on SQuAD and lower recall on DROP and ROPES with the format
instruction, and it therefore attains a clearly better precision--recall balance on
every dataset. These results show that the improvements of CLIK do not come from
longer or less constrained outputs, but from generating better-scoped answers.
Whether these answers draw on the context or on the base model's prior is examined
separately in \Cref{sec:counterfactual}.

\subsection{Reducing Generated Parameters per Context}
\label{sec:clikb}

CLIK generates a full rank-$16$ adapter on the gate, up, and down projections of all
$36$ layers, which is $21.2$M parameters per context. The analysis of the generated
adapters in \Cref{sec:lora_analysis} shows that the gate and up updates and the input
factor $\mathbf{A}^{\mathrm{down}}$ of the down projection vary little across contexts,
whereas the output factor $\mathbf{B}^{\mathrm{down}} \in \mathbb{R}^{2560 \times r}$
carries most of the context-dependent variation. We therefore evaluate \textbf{CLIK-B},
a variant in which the gate and up adapters and the down input factors
$\mathbf{A}^{\mathrm{down}}_\ell$ are ordinary trainable parameters shared across all
contexts, and the generator produces only
$\{\mathbf{B}^{\mathrm{down}}_\ell(c)\}_{\ell=1}^{L}$ from $\bar{\mathbf{H}}(c)$.
Because the shared factors are the same for every context,
$\mathbf{A}^{\mathrm{down}}_\ell$ can no longer be rotated per context, so every context
must be expressed in one learned coordinate system. The shared factors are initialized
as in standard LoRA, with random $\mathbf{A}$ and zero $\mathbf{B}$, and together with the
zero-initialized output head of the generator (\Cref{app:pg_config}) the adapter starts
from zero exactly as in CLIK. The per-context state shrinks to
$36 \times 2560 \times 16 = 1.47$M parameters ($2.8$\,MiB in BF16), a $14.4\times$
reduction. With $r \times c = 16 \times 512$ weight tokens, $\mathbf{B}^{\mathrm{down}}_\ell$
fills exactly $5$ tokens, so the generator produces $5$ instead of $72$ weight tokens per
layer ($20$ instead of $288$ generator tokens) and uses only its $\mathbf{B}$ output
head, with all other components unchanged. CLIK-B also simplifies serving: the shared
gate and up updates can be merged into the backbone weights once, and in a batch of
requests with different contexts the down projection computes
$\mathbf{W}\mathbf{x} + \frac{\alpha}{r}\mathbf{B}^{\mathrm{down}}_\ell(c_i)\,(\mathbf{A}^{\mathrm{down}}_\ell \mathbf{x})$,
where $\mathbf{A}^{\mathrm{down}}_\ell \mathbf{x}$ is a single dense matrix
multiplication for the whole batch and only the $2560 \times r$ matrix differs per
request. CLIK-B is trained with the same objective, data, and number of training steps
as CLIK.

\begin{table*}[t]
\centering
\caption{Reducing the parameters generated per context. CLIK-B generates only
$\mathbf{B}^{\mathrm{down}}$ and shares the remaining factors across contexts, under the
same backbone, data, and two evaluation protocols as \Cref{tab:main}. \# Per-context is
the number of generated parameters per context. The full adapter is shown in
\textbf{bold}.}
\label{tab:clikb}
\resizebox{\textwidth}{!}{%
\tablestyle{5pt}{1.2}
\begin{tabular}{l c ccc ccc ccc ccc}
\shline
\noalign{\vspace{1pt}}
 & & \multicolumn{6}{c}{F1 Score} & \multicolumn{6}{c}{ROUGE-L} \\
\cmidrule(lr){3-8} \cmidrule(lr){9-14}
Model & \# Per-context & \multicolumn{3}{c}{Single-Hop QA} & \multicolumn{3}{c}{Multi-Hop QA}
 & \multicolumn{3}{c}{Single-Hop QA} & \multicolumn{3}{c}{Multi-Hop QA} \\
\cmidrule(lr){3-5} \cmidrule(lr){6-8} \cmidrule(lr){9-11} \cmidrule(lr){12-14}
 & & SQuAD & DROP & ROPES & HotpotQA & MuSiQue & 2WikiMQA
      & SQuAD & DROP & ROPES & HotpotQA & MuSiQue & 2WikiMQA \\
\noalign{\vspace{1pt}}
\shline
\noalign{\vspace{1pt}}
\rowcolor{gray!10} \multicolumn{14}{c}{\textit{without Format Instruction}} \\
\noalign{\vspace{1pt}}
CLIK-B & 1.47M & 22.0 & 4.2 & 3.7 & 7.1 & 3.0 & 6.5 & 22.0 & 4.1 & 3.7 & 7.0 & 3.0 & 6.2 \\
\textbf{CLIK} & 21.2M & \textbf{24.5} & \textbf{5.3} & \textbf{4.2} & \textbf{7.9} & \textbf{3.4} & \textbf{8.2} & \textbf{24.5} & \textbf{5.3} & \textbf{4.2} & \textbf{7.6} & \textbf{3.4} & \textbf{7.4} \\
\noalign{\vspace{1pt}}
\shline
\noalign{\vspace{1pt}}
\rowcolor{gray!10} \multicolumn{14}{c}{\textit{with Format Instruction}} \\
\noalign{\vspace{1pt}}
CLIK-B & 1.47M & 44.2 & 11.1 & 46.7 & 24.0 & 8.8 & 23.4 & 44.0 & 11.1 & 46.5 & 23.8 & 8.7 & 23.2 \\
\textbf{CLIK} & 21.2M & \textbf{46.3} & \textbf{14.3} & \textbf{49.4} & \textbf{28.7} & \textbf{12.1} & \textbf{29.2} & \textbf{45.1} & \textbf{14.3} & \textbf{48.8} & \textbf{28.5} & \textbf{12.1} & \textbf{28.8} \\
\noalign{\vspace{1pt}}
\shline
\end{tabular}%
}
\end{table*}

\Cref{tab:clikb} shows that CLIK-B retains most of the performance of CLIK. With the
format instruction, it stays behind only CLIK and ahead of every baseline in
\Cref{tab:main} on every dataset, and it stays closest to CLIK on SQuAD and ROPES.
The loss is larger on DROP and on the multi-hop datasets, where answering requires
combining several facts from the context and therefore relies more on the
context-dependent part of the factors that CLIK-B shares. Without the format
instruction, CLIK-B is comparable to the strongest baselines. This makes CLIK-B an
attractive trade-off when per-context storage matters, since its per-context state is
more than an order of magnitude smaller than that of CLIK.

\subsection{Controlled Comparison at Matched Scale}
\label{sec:controlled}

The baselines in \Cref{tab:main} share the backbone, training data, and generated
adapter with CLIK, but their generators differ from that of CLIK not only in
architecture but also in size, so \Cref{tab:main} alone does not separate the
contribution of the architecture from that of scale. We therefore vary only the scale
of the generators and keep everything else as in \Cref{tab:main}: the same
Qwen3-4B-Instruct backbone, the same FineWeb-Edu question--answer pairs, rank $16$
adapters on the gate, up, and down projections of all $36$ layers, the same number of
training steps, and the training objective of each method, i.e., distillation for
Doc-to-LoRA and CLIK and next-token prediction for SHINE (details in
\Cref{app:baselines}). In \Cref{tab:controlled}, D2L and SHINE denote these native-size
models, which are therefore those of \Cref{tab:main}.
To separate architecture from capacity, we scale each baseline to the parameter count
of CLIK, denoted D2L-L and SHINE-L, and we evaluate a small CLIK-S that reduces the
generator from $12$ to $6$ blocks and keeps everything else unchanged, which halves its
parameter count and leaves it smaller than D2L.
D2L-L widens the latent dimension of Doc-to-LoRA from $512$ to $576$, and SHINE-L
deepens the M2P Transformer from $4$ to $14$ layers.
\Cref{tab:controlled} reports generator size, per-context adapter size, training
compute, and F1 with the format instruction. Matching parameters does not match
compute: at the same number of steps CLIK needs several times the GPU-hours of
Doc-to-LoRA, because its generator applies all of its parameters to every weight token
of the adapter for every context, whereas Doc-to-LoRA concentrates most of its
parameters in an output head that is applied to only a few queries per layer, and the
M2P Transformer of SHINE is much smaller. Halving the depth, CLIK-S also
halves the generator compute of CLIK.

Scaling the baselines to the size of CLIK does not close the gap. D2L-L remains
essentially unchanged relative to D2L, and SHINE-L improves on every
dataset but stays clearly behind CLIK at the same parameter count. In the other
direction, CLIK-S, with half the parameters of CLIK, outperforms both the larger
D2L and SHINE of comparable size on every dataset, so the advantage
of CLIK comes from its architecture rather than from its size. CLIK-B (\Cref{sec:clikb}) further shows that
this advantage does not depend on a large training budget: it requires the least
training compute of all generators and still achieves a higher average F1 than every
baseline, including D2L-L and SHINE-L.

\begin{table*}[t]
    \centering
    \caption{Controlled comparison of generator scale under matched backbone
    (Qwen3-4B-Instruct), training data (FineWeb-Edu), adapter rank ($16$), target modules
    (gate, up, and down projections of all $36$ layers), and number of training steps. Each
    method keeps its training objective. D2L and SHINE are Doc-to-LoRA and SHINE at native
    size, the models of \Cref{tab:main}; D2L-L and SHINE-L enlarge the generator until it
    matches the parameter count of CLIK. \# Gen.\ is the number of generator parameters and
    \# Per-context the number of generated parameters per context. We report training
    GPU-hours (\Cref{app:cost}) and F1 with the format instruction.}
    \label{tab:controlled}
    \tablestyle{4pt}{1.2}
    \begin{tabular}{l ccc ccc ccc c}
    \shline
    \noalign{\vspace{1pt}}
    Method & \# Gen. & \# Per-context & GPU-hours & SQuAD & DROP & ROPES & HotpotQA & MuSiQue & 2WikiMQA & Avg.\ F1 \\
    \noalign{\vspace{1pt}}
    \shline
    \noalign{\vspace{1pt}}
    D2L    & 0.89B & 21.2M & 13.0k & 29.6 & 7.8 & 12.1 & 15.3 & 6.4 & 10.2 & 13.6 \\
    D2L-L        & 1.01B & 21.2M & 13.1k & 29.2 & 8.1 & 12.4 & 15.1 & 6.8 & 10.5 & 13.7 \\
    SHINE          & 0.48B & 21.2M & 13.6k & 35.2 & 10.4 & 34.4 & 22.2 & 6.0 & 18.7 & 21.2 \\
    SHINE-L               & 1.00B & 21.2M & 23.7k & 39.8 & 12.7 & 40.5 & 25.1 & 9.8 & 23.9 & 25.3 \\
    \hline
    \noalign{\vspace{1pt}}
    CLIK-S             & 0.51B & 21.2M & 23.5k & 35.4 & 11.9 & 39.9 & 24.8 & 8.3 & 21.7 & 23.7 \\
    CLIK-B                       & 1.01B & 1.47M & 10.3k & 44.2 & 11.1 & 46.7 & 24.0 & 8.8 & 23.4 & 26.4 \\
    CLIK                         & 1.01B & 21.2M & 39.1k & 46.3 & 14.3 & 49.4 & 28.7 & 12.1 & 29.2 & 30.0 \\
    \noalign{\vspace{1pt}}
    \shline
    \end{tabular}%
    \end{table*}

\subsection{Context Following of the Generated Adapter}
\label{sec:counterfactual}

A high score on SQuAD or HotpotQA does not by itself show that the generator internalizes
the given context. The generator could instead have adapted to the evaluation
distribution, and the base model already answers part of these questions from its prior
(e.g., on ROPES, where Base is already competitive without any context). We address the
first concern with in-domain training data and the second with counterfactual probes.

\paragraph{In-domain training data.} The question--answer pairs synthesized from
FineWeb-Edu have long, free-form answers (\Cref{sec:data}), whereas the evaluation
datasets have short reference answers, so every evaluation set is out of distribution
for the generator. To measure how much in-domain supervision changes the results, we
follow the data ablation of Doc-to-LoRA~\citep{charakorn2026doc} and add the passages
and questions of the training splits of SQuAD, DROP, and ROPES to the training data,
keeping the distillation objective and all other settings unchanged
(\Cref{tab:indomain}). In-domain data yields only a modest gain. The gain is largest on
DROP, whose numerical and multi-span answers differ most from the free-form answers of
FineWeb-Edu, which matches the observation of~\citet{charakorn2026doc}, and it is small
on SQuAD and ROPES, whose extractive and short reasoning answers the generator already
handles well. The results of CLIK therefore do not rely on training data that matches
the evaluation distribution: training on the general-domain corpus alone already
transfers to the evaluation tasks, and in-domain supervision mainly adapts the answer
format of the task.

\begin{table}[t]
\centering
\tablestyle{5pt}{1.2}
\caption{Effect of adding the training splits of SQuAD, DROP, and ROPES to the
FineWeb-Edu training data of CLIK. F1 with the format instruction; all other settings
are unchanged.}
\label{tab:indomain}
\begin{tabular}{@{}l ccc c@{}}
\shline
\noalign{\vspace{1pt}}
Training data & SQuAD & DROP & ROPES & Avg. \\
\noalign{\vspace{1pt}}
\shline
\noalign{\vspace{1pt}}
FineWeb-Edu                               & 46.3 & 14.3 & 49.4 & 36.7 \\
FineWeb-Edu + SQuAD, DROP, ROPES          & 47.0 & 16.3 & 49.5 & 37.6 \\
\noalign{\vspace{1pt}}
\shline
\end{tabular}
\end{table}

\begin{table*}[t]
\centering
\caption{Counterfactual probes of context following (accuracy in \%, with the format
instruction). For \emph{Leak} and \emph{Prior}, lower is better; for the wrong-context
swap, the ideal accuracy equals that of Base.}
\label{tab:counterfactual}
\resizebox{\textwidth}{!}{%
\tablestyle{5pt}{1.2}
\begin{tabular}{l ccc c cc cc c}
\shline
\noalign{\vspace{1pt}}
 & \multicolumn{3}{c}{Paired counterfactual} & Novel facts & \multicolumn{2}{c}{Prior conflict} & \multicolumn{2}{c}{Wrong-context swap} & Unrelated queries \\
\cmidrule(lr){2-4} \cmidrule(lr){5-5} \cmidrule(lr){6-7} \cmidrule(lr){8-9} \cmidrule(lr){10-10}
Method & Own $\uparrow$ & Both $\uparrow$ & Leak $\downarrow$ & Acc $\uparrow$ & Ctx $\uparrow$ & Prior $\downarrow$ & Acc & Leak $\downarrow$ & Retention $\uparrow$ \\
\noalign{\vspace{1pt}}
\shline
\noalign{\vspace{1pt}}
Base            & 0.4 & 0.0 & 0.4 & 1.3 & 0.0 & 100.0 & 18.4 & 0.0 & 100.0 \\
Base + Context  & 93.8 & 88.6 & 0.2 & 91.4 & 84.7 & 11.9 & 17.1 & 6.3 & 97.8 \\
\hline
\noalign{\vspace{1pt}}
Doc-to-LoRA     & 43.2 & 22.8 & 0.6 & 44.9 & 37.9 & 55.4 & 16.5 & 5.1 & 95.6 \\
SHINE           & 53.2 & 34.9 & 0.8 & 54.6 & 39.0 & 43.0 & 17.4 & 5.5 & 95.9 \\
CLIK            & 72.5 & 56.3 & 0.5 & 67.4 & 56.1 & 32.4 & 17.5 & 5.4 & 96.4 \\
\noalign{\vspace{1pt}}
\shline
\end{tabular}%
}
\end{table*}

\paragraph{Counterfactual probes.} Neither benchmark scores nor in-domain training can
separate the internalized context from knowledge that the backbone already holds, so we
construct probes whose correct answer can only come
from the given context (construction details in \Cref{app:probes}).
\begin{itemize}
    \item \emph{Paired counterfactual documents.} Two copies of a passage differ in a
    single fact, which is replaced by a nonce value (e.g., a made-up name or date), and
    both are queried with the identical question. We report the accuracy with respect to
    each adapter's own context (\emph{Own}), the fraction of pairs where both answers are
    correct (\emph{Both}), and the fraction of answers that give the other document's
    value (\emph{Leak}).
    \item \emph{Novel facts.} Synthetic passages about fictional entities that cannot
    occur in any pretraining corpus.
    \item \emph{Strong prior conflict.} Passages that contradict a well-known fact
    (e.g., relocating a famous landmark); we report how often the answer follows the
    context (\emph{Ctx}) and how often it follows the prior (\emph{Prior}).
    \item \emph{Wrong-context swap.} The question is asked under the adapter generated
    from an unrelated passage, which carries no answer, so the ideal accuracy is that
    of Base, and we also report how often the answer uses content from the
    unrelated passage (\emph{Leak}).
    \item \emph{Unrelated-query interference.} General-knowledge questions unrelated to
    the passage are asked under the adapter; we report accuracy relative to Base
    (\emph{Retention}), which measures collateral damage to the backbone.
\end{itemize}
\Cref{tab:counterfactual} shows that CLIK follows the internalized context rather than
the prior of the backbone. Base stays at chance level on the paired counterfactual
documents and on the novel facts, which confirms that these probes can only be solved
from the context. The probes that require writing the context into the adapter
reproduce the ranking of \Cref{tab:main}. On paired counterfactual documents, each
adapter of CLIK answers with the nonce value of its own copy, and CLIK answers both
copies of a pair correctly far more often than Doc-to-LoRA and SHINE, so changing a
single fact in the context changes the answer accordingly. On novel facts, CLIK
recovers the answer from the adapter alone with the highest accuracy among the
generators. Under strong prior conflict, where Base always gives the original fact,
CLIK follows the edited context more often than the prior, whereas Doc-to-LoRA and
SHINE follow the prior more often than the context.

The probes of specificity and collateral effects behave alike for all three
generators, since all of them are trained to internalize the facts of a given context.
Their adapters almost never give the value of the other copy of a pair, stay close to
the accuracy of Base under a wrong-context swap and inject less content from the
unrelated passage than Base + Context, and preserve the general knowledge of the
backbone on unrelated queries. The stronger internalization of CLIK therefore comes
without any loss of specificity: its adapter carries the specific facts of its context
and leaves the remaining behavior of the backbone intact.

\begin{figure}[t]
    \centering
    \includegraphics[width=0.7\linewidth]{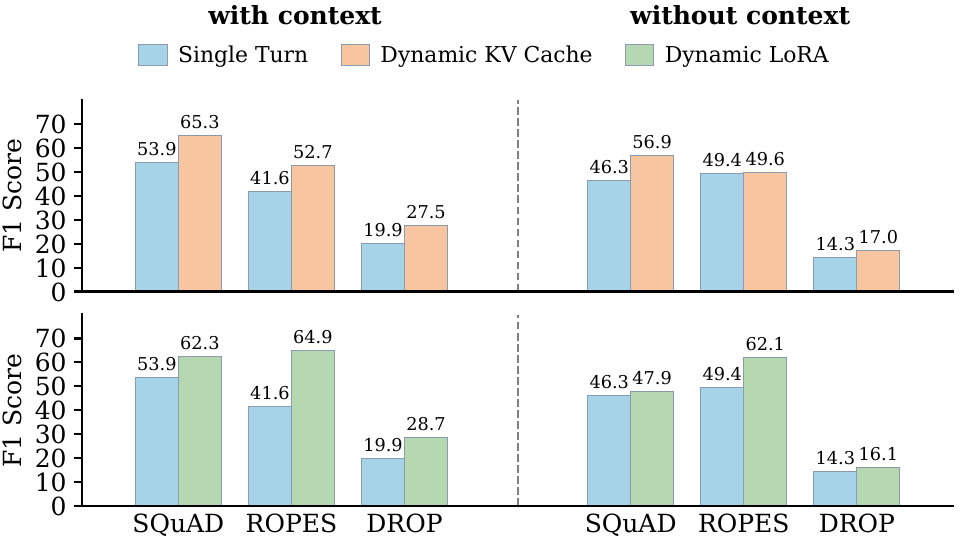}
    \caption{Multi-turn evaluation of CLIK on the SQuAD, ROPES, and DROP
    datasets, comparing the single-turn setting against two multi-turn scenarios
    that differ in how conversational state is propagated. \emph{Top (Dynamic KV
    cache)}: the base model accumulates a KV cache across turns while the LoRA
    generated from the initial context is kept fixed. \emph{Bottom (Dynamic
    LoRA)}: the KV cache is discarded and CLIK regenerates the LoRA parameters at
    each turn from the updated context. Both scenarios improve over the
    single-turn baseline without any multi-turn-specific training, but neither has a
    per-turn cost that is independent of the history (\Cref{tab:multiturn_cost}).}
    \label{fig:context_recovery}
\end{figure}

\subsection{Multi-turn Evaluation}
\label{sec:multiturn}
Beyond the single-turn QA task examined above, we study the behavior of CLIK in
realistic multi-turn conversations to assess its effectiveness under conditions that
more closely resemble practical deployment. To construct such a setting from
existing resources, we use the fact that each context in the QA datasets is
associated with multiple questions, which we issue sequentially to form a multi-turn
dialogue grounded in a shared initial context~\citep{bai2024mt}. Within this setup,
we consider two complementary scenarios that differ in how information from
preceding turns is propagated, namely \emph{dynamic KV cache} and \emph{dynamic
LoRA}, and we additionally report an exploratory attempt at a \emph{recurrent update}
whose per-turn cost does not grow with the history.

\paragraph{Dynamic KV cache.}
In this scenario, the base model maintains a KV cache that accumulates across
dialogue turns, while CLIK is responsible only for internalizing the initial
context $c_0$ into LoRA parameters. At the first turn, the base model is conditioned
on the question $q_1$ alone and produces the answer $a_1$ under the generated LoRA.
At the second turn, the pair $(q_1, a_1)$ is provided together with $q_2$, while the
LoRA parameters remain fixed. As shown in \Cref{fig:context_recovery}, this
scheme yields consistent gains over the single-turn setting. We attribute these
gains to an emergent transition toward in-context learning, in which CLIK first
enables the model to produce a correct answer and the accumulated dialogue history
then acts as reliable in-context evidence that improves the quality of subsequent
responses. This effect arises without any fine-tuning on multi-turn data, which
confirms the robustness of CLIK with respect to the base model's own outputs. In this
scenario the context is encoded only once, but the backbone prompt grows with the
dialogue history.

\paragraph{Dynamic LoRA.}
In this scenario, we further dispense with the KV cache entirely and instead allow
CLIK to regenerate LoRA parameters dynamically as the conversation unfolds. The
first turn is identical to that of the dynamic KV cache scenario. At the second
turn, however, the initial context $c_0$ and the pair $(q_1, a_1)$ are recombined
into an updated context $c_1$ and passed in full through the parameter generator to
synthesize new LoRA parameters, while the base model is still conditioned only on
$q_2$. As reported in \Cref{fig:context_recovery}, this scheme likewise
surpasses the single-turn setting and achieves the strongest results under several
configurations. These results show that CLIK encodes the behavioral regularities
that emerge during multi-turn interaction, such as the output-format constraints
induced by the format instruction, directly into the LoRA parameters, and adapts them
dynamically to refine its behavior over successive turns. This scheme is, however,
not ``encode once'': the encoder and generator re-read the original context plus the
entire history at every turn, so their input, and hence the per-turn cost, grows
with the number of turns (\Cref{tab:multiturn_cost}).

In summary, the advantages of CLIK extend from the single-turn setting to
multi-turn dialogue under two distinct mechanisms for propagating conversational
state. Whether preceding turns are carried forward through the KV cache or
re-internalized into newly generated parameters, CLIK consistently improves upon the
single-turn baseline without any multi-turn-specific training. Neither mechanism,
however, has a per-turn cost that is independent of the history, so the
``encode once'' amortization of CLIK applies to repeated queries against a fixed
document rather than to a dialogue whose content keeps changing.

\paragraph{An exploratory attempt: recurrent update.} To obtain a per-turn cost that
is independent of the history, we explored a recurrent update in which the adapter of
turn $t$ is produced from the previous adapter and the new turn
$x_t = (q_t, \hat{\mathbf{a}}_t)$ only,
\begin{equation}
    \Phi_t \;=\; G_\theta\!\bigl(\bar{\mathbf{H}}(x_t);\, \Phi_{t-1}\bigr), \qquad \Phi_0 = G_\theta\!\bigl(\bar{\mathbf{H}}(c_0)\bigr).
    \label{eq:recurrent}
\end{equation}
In CLIK the generator tokens are initialized from learned layer and within-layer
position embeddings (\Cref{sec:generator}). The recurrent update replaces this
initialization with the weight tokens of the previous adapter: $\Phi_{t-1}$ is tokenized
as in \Cref{sec:generator}, which gives exactly the shape of the position embeddings,
and the resulting tokens serve as the initial generator tokens, while the
cross-attention reads only the hidden states of the new turn. The encoder and
generator then process only the $|x_t|$ tokens of the new turn, regardless of the
length of the dialogue, and the previous turns reach the new adapter only through
$\Phi_{t-1}$. The update is applied to the trained CLIK generator as is, without any
additional training.

As shown in \Cref{tab:recurrent}, this attempt was not successful: the recurrent update
is clearly worse than the single-turn setting, let alone dynamic LoRA. We attribute
this to the gap between the two kinds of condition. CLIK learns a text$\to$weight
mapping in which the context enters only through cross-attention to the backbone's own
hidden states, a representation that is already semantically organized, while the
initial tokens are a fixed, context-independent starting point that tells each token
which layer and which position it will write. Taking $\Phi_{t-1}$ as a condition
instead requires a weight$\to$weight mapping that the generator has never learned.
The weight tokens of $\Phi_{t-1}$ are consecutive chunks of flattened factor matrices
rather than semantic units, they carry no explicit layer or position identity, and
the same update $\Delta\mathbf{W} = \mathbf{B}\mathbf{A}$ can be written by any factor
pair $(\mathbf{B}\mathbf{R}, \mathbf{R}^{-1}\mathbf{A})$, so they are not a canonical
encoding of what has been internalized. Replacing the position embeddings with them
therefore removes the starting point the generator relies on without supplying a
condition it can read, which degrades even the internalization of the new turn and
explains why the result falls below the single-turn setting. The existing
text$\to$condition pathway thus does not transfer directly to a weight$\to$condition
one without training. Learning both jointly, i.e., a multi-task generator trained to
internalize a context from text and to update an existing adapter from new turns,
for example with a dedicated input path for the previous adapter or with a
gauge-invariant encoding or a shared coordinate system as in CLIK-B, in which adapters
of different turns are directly comparable, is left to future work.

\begin{table}[t]
\centering
\tablestyle{5pt}{1.2}
\caption{Exploratory recurrent update (\Cref{eq:recurrent}) compared with the
multi-turn schemes above. ``Generator input'' is the number of context tokens read by
the encoder and generator at turn $t$. F1 is averaged over turns $1$--$5$ with the
format instruction.}
\label{tab:recurrent}
\begin{tabular}{@{}l c ccc@{}}
\shline
\noalign{\vspace{1pt}}
Scenario & Generator input & SQuAD & ROPES & DROP \\
\noalign{\vspace{1pt}}
\shline
\noalign{\vspace{1pt}}
Single turn (no history) & $0$ for $t>1$ & 46.3 & 49.4 & 14.3 \\
Dynamic LoRA & $|c_0| + \sum_{i<t}|x_i|$ & 47.9 & 62.1 & 16.1 \\
\hline
\noalign{\vspace{1pt}}
Recurrent update & $|x_{t-1}|$ & 25.2 & 31.1 & 7.4 \\
\noalign{\vspace{1pt}}
\shline
\end{tabular}
\end{table}

\begin{table*}[t]
\centering
\caption{Per-turn cost and quality of multi-turn scenarios. $|c_0|$ is the length of the
initial context, $|x_i| = |q_i| + |a_i|$ the length of turn $i$. ``Encoder / generator
input'' and ``Backbone prompt'' give the number of tokens processed at turn $t$. Latency is
the end-to-end time of turn $5$ for $|c_0|=2048$
(\Cref{app:cost}); F1 is averaged over turns $1$--$5$ with the format
instruction.}
\label{tab:multiturn_cost}
\resizebox{\textwidth}{!}{%
\tablestyle{5pt}{1.2}
\begin{tabular}{l c c c ccc}
\shline
\noalign{\vspace{1pt}}
Scenario & Encoder / generator input at turn $t$ & Backbone prompt at turn $t$ & Latency at $t{=}5$ (ms) & SQuAD & ROPES & DROP \\
\noalign{\vspace{1pt}}
\shline
\noalign{\vspace{1pt}}
Single turn (no history)             & $0$ for $t>1$ & $|q_t|$ & 109 & 46.3 & 49.4 & 14.3 \\
Dynamic KV cache                     & $0$ for $t>1$ & $|q_t| + \sum_{i<t}|x_i|$ & 109 & 56.9 & 49.6 & 17.0 \\
Dynamic LoRA                         & $|c_0| + \sum_{i<t}|x_i|$ & $|q_t|$ & 564 & 47.9 & 62.1 & 16.1 \\
\noalign{\vspace{1pt}}
\shline
\end{tabular}%
}
\end{table*}

\subsection{Cost and Efficiency}
\label{sec:efficiency}

Replacing a context by an adapter only pays off if the adapter is cheaper to keep and
serve than the context. We compare against the strongest context-based alternatives,
namely prefix caching, which prefills the context once and reuses its KV cache for
every query, KV-cache compression, and prompt compression, as well as Doc-to-LoRA and
SHINE.

\paragraph{Per-context state.} \Cref{tab:storage} compares the state that must be
stored per context. With the adapter, the backbone answers from the query alone, so
inference no longer needs the KV cache of the context, and the adapter takes its place
as the per-context state. The rank-$16$ CLIK adapter is as large as the KV cache of a
context of a few hundred tokens, so for the long documents that context internalization
targets it is several times smaller than the KV cache it removes, and only for short
contexts is it the larger of the two; in every case it has to be loaded and switched
per request. CLIK-B shrinks the per-context state by more than an order of magnitude,
to the size of the KV cache of only tens of tokens.

\begin{table}[t]
\centering
\tablestyle{5pt}{1.2}
\caption{Per-context state in BF16 for Qwen3-4B, where $n$ is the number of context
tokens. ``KV tokens'' expresses each size as the number of tokens whose full KV cache
occupies the same memory ($144$\,KiB per token).}
\label{tab:storage}
\begin{tabular}{@{}l r r r@{}}
\shline
\noalign{\vspace{1pt}}
Per-context state & Parameters & Size (MiB) & KV tokens \\
\noalign{\vspace{1pt}}
\shline
\noalign{\vspace{1pt}}
Full KV cache, $n=512$   & 37.7M  & 72.0  & 512 \\
Full KV cache, $n=2048$  & 151.0M & 288.0 & 2048 \\
Compressed KV cache, $256$-token budget & 18.9M & 36.0 & 256 \\
\hline
\noalign{\vspace{1pt}}
Rank-$16$ FFN LoRA (Doc-to-LoRA, SHINE, CLIK) & 21.2M & 40.5 & 288 \\
CLIK-B ($\mathbf{B}^{\mathrm{down}}$ only)        & 1.47M & 2.8  & 20 \\
\noalign{\vspace{1pt}}
\shline
\end{tabular}
\end{table}

\paragraph{Latency, memory, and throughput.} \Cref{tab:efficiency} reports inference
costs for contexts of $2048$ tokens (setup in \Cref{app:cost}). \emph{Inter. (Internalization)} is the one-time cost per
context (prefill for context-based methods, encoder plus generator pass for
adapter-based methods), \emph{TTFT} is the per-query time to first token after
internalization, \emph{Switch} is the time to load a stored per-context state from host
memory, \emph{TP. (Throughput)} is the decoding throughput for a batch of $32$ requests
with $32$ distinct contexts, using segmented multi-adapter kernels~\citep{chen2024punica,sheng2024slora}
for the adapter-based methods, and \emph{NQ. (New question)} is the added latency of one further
question on the same document: the TTFT when the internalized state is reused, and the
internalization time plus the TTFT when SnapKV selects the cache again. \emph{Memory} refers to this batched run
and includes the backbone, any generator or compressor, the $32$ per-context states, and
the transient memory of one internalization.

Internalization is more expensive for CLIK than prefilling the context, because the
generator adds a pass over all weight tokens of the adapter, whereas for CLIK-B this
extra cost is small. After internalization, the time to first token of the
adapter-based methods is close to that of prefix caching, since for a single request
both are dominated by reading the backbone weights. The advantages of the adapter lie
in memory: because the backbone no longer needs the KV cache of the context, each
request holds only its adapter and the KV cache of its own query and answer. The
adapter is several times smaller than the context KV cache it removes and
correspondingly faster to load, and a batch of requests with distinct contexts decodes
markedly faster with lower peak memory. SnapKV reaches a similar throughput by reading
its retained KV cache, and the internalization time counts that selection once. A further
question selects the cache again, so its added latency is that internalization plus the
TTFT. An adapter is generated once for the document and then reused, so a further question
costs only the TTFT. LLMLingua-2 compresses the context once and pays its TTFT on every
query, which already includes prefilling the compressed prompt.

\begin{table*}[t]
\centering
\caption{Inference cost for $2048$-token contexts (\Cref{app:cost}). \emph{New question} is the added latency of one further question on the same document, taken from the other columns: the TTFT when the internalized state is reused, and the internalization time plus the TTFT for SnapKV, which selects its cache again. Avg.\ F1 is over the six datasets with the format
instruction.}
\label{tab:efficiency}
\resizebox{\textwidth}{!}{%
\tablestyle{5pt}{1.2}
\begin{tabular}{l c c c c c c c c}
\shline
\noalign{\vspace{1pt}}
Method & Inter. (ms) & PCS. (MiB) & Switch (ms) & TTFT (ms) & TP. (tok/s) & Memory (GB) & NQ (ms) & Avg.\ F1 \\
\noalign{\vspace{1pt}}
\shline
\noalign{\vspace{1pt}}
Base + Context         & --    & --    & --   & 185.2 & 5.2k & 19.2 & 185.2 & 50.4 \\
Base + Context (prefix)   & 182.1 & 288.0 & 6.1  & 3.7   & 5.2k & 19.2 & 3.7   & 50.4 \\
SnapKV ($256$-token)     & 182.6 & 36.0  & 0.8  & 3.3   & 9.0k & 10.9 & 185.9 & 39.4 \\
LLMLingua-2 ($4\times$)         & 25.1  & $<$0.01 & $<$0.1 & 46.1 & 8.1k & 13.1 & 46.1  & 32.9 \\
\hline
\noalign{\vspace{1pt}}
Doc-to-LoRA                     & 282.2 & 40.5  & 0.9  & 3.5   & 8.3k & 15.9 & 3.5   & 13.6 \\
SHINE                           & 258.7 & 40.5  & 0.9  & 3.5   & 8.3k & 12.1 & 3.5   & 21.2 \\
CLIK-B                          & 204.4 & 2.8   & 0.1  & 3.3   & 8.9k & 12.0 & 3.3   & 26.4 \\
CLIK                            & 427.0 & 40.5  & 0.9  & 3.5   & 8.3k & 13.6 & 3.5   & 30.0 \\
\noalign{\vspace{1pt}}
\shline
\end{tabular}%
}
\end{table*}

\subsection{Ablation Study on Hyper-parameters}

\begin{table*}[t]
    \centering
    \tablestyle{5pt}{1.2}
    \caption{Ablation studies on SQuAD, reported as Precision / Recall / F1 (P / R /
    F1). We vary three factors independently: the hidden state read by the
    cross-attention, the peak learning rate, and the training corpus. The
    configuration adopted in our final model is highlighted in \textbf{bold}.}
    \label{tab:ablation}
    \begin{tabular}{@{}lccc@{\hspace{2em}}lccc@{\hspace{2em}}lccc@{}}
        \shline
        \noalign{\vspace{1pt}}
        \multicolumn{4}{c}{\textit{Hidden state}} &
        \multicolumn{4}{c}{\textit{Learning rate}} &
        \multicolumn{4}{c}{\textit{Training dataset}} \\
        \cmidrule(r){1-4} \cmidrule(lr){5-8} \cmidrule(l){9-12}
        & P & R & F1 & & P & R & F1 & & P & R & F1 \\
        \noalign{\vspace{1pt}}
        \shline
        \noalign{\vspace{1pt}}
        Last layer   & 12.5 & 78.0 & 14.8 & $1\mathrm{e}{-}6$ & 18.8 & 72.7 & 30.4 & SQuAD & 21.5 & 20.5 & 20.9 \\
        Middle layer & 39.0 & 78.2 & 45.2 & $1\mathrm{e}{-}5$ & 28.5 & 78.8 & 41.2 & MS MARCO       & 32.8 & 58.6 & 39.5 \\
        Mean         & 36.0 & 79.5 & 44.3 & $1\mathrm{e}{-}4$ & 38.5 & \textbf{81.9} & 45.2 & TransMLA    & 32.6 & 70.2 & 44.8 \\
        \rowcolor{gray!10} ELMo         & \textbf{39.4} & \textbf{81.7} & \textbf{46.3} & $5\mathrm{e}{-}5$ & \textbf{39.4} & 81.7 & \textbf{46.3} & FineWeb-Edu    & \textbf{39.4} & \textbf{81.7} & \textbf{46.3} \\
        \noalign{\vspace{1pt}}
        \shline
    \end{tabular}
\end{table*}

\paragraph{Hidden state for cross-attention.}
We first ablate which hidden representation the cross-attention reads from when
generating the LoRA. Relying on a single fixed layer is suboptimal, as the
last-layer representation performs poorly, and although the middle layer and a
uniform mean over layers are clearly better, both still remain below the best
configuration. Replacing the fixed choice with an ELMo-style learnable mixture of
the hidden states from all layers performs best, which shows that
the most useful contextual signal is distributed across depth and that letting the
model learn how to combine layers is preferable to committing to any single one.

\paragraph{Learning rate.}
The learning rate controls how quickly the generator converges. A very small rate
underfits within our training budget, and performance improves as the rate
increases, peaking at an intermediate value. A larger rate speeds up convergence, but
increasing it further no longer helps and slightly degrades the result, which
reflects the instability that accompanies an overly aggressive step size. We
therefore adopt the best-performing rate, which converges quickly while remaining
stable.

\paragraph{Training dataset.}
Finally, we vary the corpus used to train the generator. A short-answer dataset
such as SQuAD yields the lowest F1, with a particularly low recall, even
though it shares the answer format of the evaluation, while larger and more general
sources help progressively. The best result comes from FineWeb-Edu,
showing that increasing the scale and diversity of the training data, rather than
matching the answer format of the evaluation, is what truly drives performance.
This holds even though the QA pairs derived from FineWeb-Edu contain long answers.
We deliberately avoid training on a fixed short-answer dataset like SQuAD and do not
impose any constraint on the base model's outputs, because our goal is to compress
and preserve the base model's native ability as faithfully as possible rather than
to specialize it toward a particular response style.

\subsection{Ablation of the Generator Architecture}
\label{sec:arch_ablation}

\Cref{tab:ablation} studies input and training choices, but the component that
distinguishes CLIK from prior generators is the interaction among weight tokens. We
therefore ablate it (\Cref{tab:arch_ablation}): \emph{across-layer attention} and
\emph{within-layer attention} are removed individually and jointly. Removing both
turns CLIK into a generator whose weight tokens interact only through the shared
context, i.e., independent per-fragment generation as in prior work. To keep capacity
comparable, we adjust the feed-forward width of every variant so that its parameter
count stays within $5\%$ of the full model ($12{,}288$ when one attention sublayer per
block is removed and $16{,}384$ when both are removed).

Both forms of interaction contribute to the performance of CLIK. Removing across-layer
attention lowers every metric, and removing within-layer attention causes a much larger
drop, which shows that the weight tokens of one layer must be generated in coordination
to form a usable adapter. Removing both is far more harmful than removing either one
alone, so the two sublayers complement each other. The resulting independent-fragment
generator falls to the level of Doc-to-LoRA in \Cref{tab:main}, even though it keeps the
cross-attention and the parameter count of CLIK. The loss is concentrated in precision
and is largest on HotpotQA, whose questions require combining several facts of the
context. The interaction among weight tokens is therefore the key architectural
component of CLIK.

\begin{table}[t]
\centering
\tablestyle{5pt}{1.2}
\caption{Ablation of the weight-token interaction in the generator. All variants are trained under the
same data and number of steps and are parameter-matched to within $5\%$. We report P / R / F1
on SQuAD, F1 on HotpotQA, and the average F1 over the six datasets, all with the format
instruction.}
\label{tab:arch_ablation}
\begin{tabular}{@{}l c ccc c c@{}}
\shline
\noalign{\vspace{1pt}}
 & & \multicolumn{3}{c}{SQuAD} & HotpotQA & All \\
\cmidrule(lr){3-5} \cmidrule(lr){6-6} \cmidrule(l){7-7}
Variant & Gen.\ params & P & R & F1 & F1 & Avg.\ F1 \\
\noalign{\vspace{1pt}}
\shline
\noalign{\vspace{1pt}}
\rowcolor{gray!10} CLIK (full) & 1.01B & 39.4 & 81.7 & 46.3 & 28.7 & 30.0 \\
\hline
\noalign{\vspace{1pt}}
w/o across-layer attention & 1.01B & 36.3 & 79.4 & 44.2 & 26.8 & 27.6 \\
w/o within-layer attention & 1.01B & 31.1 & 72.8 & 38.7 & 20.2 & 19.8 \\
w/o both (independent fragments) & 1.01B & 22.4 & 59.7 & 32.6 & 10.8 & 11.4 \\
\noalign{\vspace{1pt}}
\shline
\end{tabular}
\end{table}

\begin{figure}[t]
    \centering
    \includegraphics[width=0.7\linewidth]{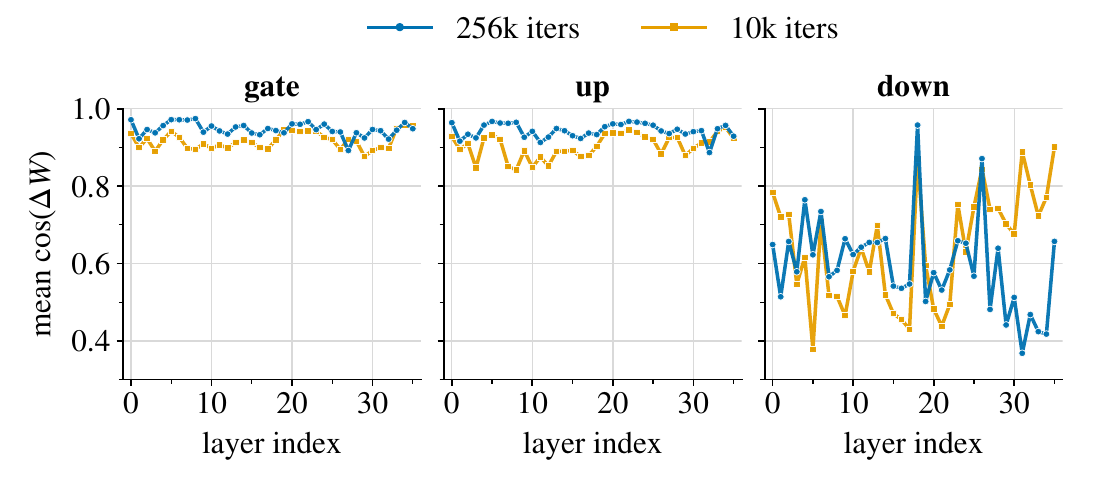}
    \caption{Per-layer mean cross-context $\cos(\Delta W)$ for the gate, up, and
    down projections, comparing an early (10k iters) and a late (256k iters)
    checkpoint of CLIK. Higher values denote updates that are more consistent across
    different contexts.}
    \label{fig:lora_per_layer}
\end{figure}
\begin{figure}[!t]
    \centering
    \begin{subfigure}[t]{0.32\linewidth}
        \centering
        \includegraphics[width=\linewidth]{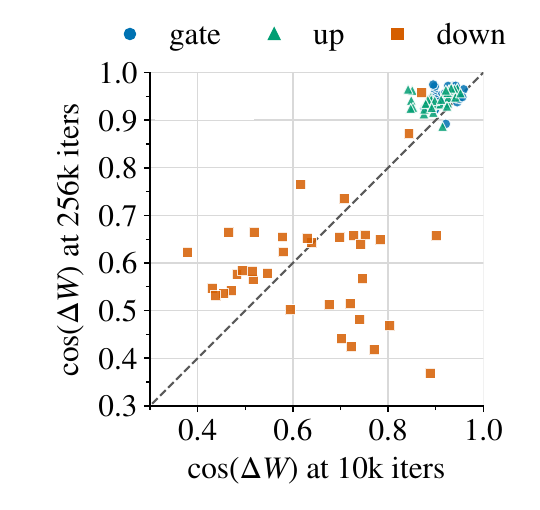}
        \caption{Per-module $\cos(\Delta W)$ at the early versus late
        checkpoint, split by module kind.}
        \label{fig:lora_scatter}
    \end{subfigure}%
    \hfill
    \begin{subfigure}[t]{0.32\linewidth}
        \centering
        \includegraphics[width=\linewidth]{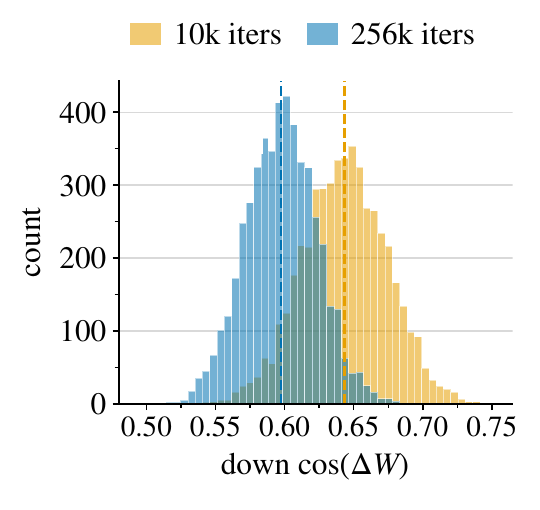}
        \caption{Off-diagonal $\cos(\Delta W)$ for the down projection over
        all context pairs, early versus late.}
        \label{fig:lora_down_hist}
    \end{subfigure}%
    \hfill
    \begin{subfigure}[t]{0.32\linewidth}
        \centering
        \includegraphics[width=\linewidth]{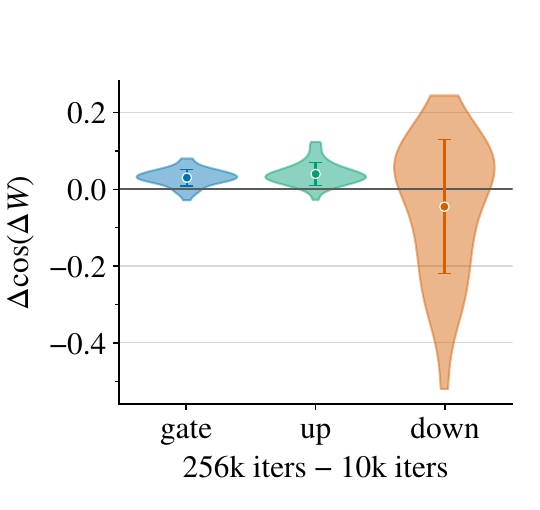}
        \caption{Per-kind change in $\cos(\Delta W)$ from early to late
        (mean $\pm$ std).}
        \label{fig:lora_violin}
    \end{subfigure}
    \caption{Distributional view of the generated update $\Delta W=(\alpha/r)BA$,
    comparing the early and late checkpoints. The cross-context cosine
    $\cos(\Delta W)$ measures how consistent the update is across contexts: high
    values indicate a shared, context-independent component, while low values
    indicate context-specific adaptation. Because it is computed on the product
    $BA$, it is invariant to the reparameterization $(B, A) \mapsto (BR, R^{-1}A)$.}
    \label{fig:lora_dist}
\end{figure}

\subsection{Analysis of LoRA Patterns}
\label{sec:lora_analysis}

\paragraph{Training Dynamics.} To understand what CLIK learns, we compare the
adapters it generates at an early and a late training checkpoint, analyzing the effective update $\Delta W=(\alpha/r)BA$ over a set of
contexts. Two trends are evident. First, training increases the magnitude of the
adaptation: the mean Frobenius norm $\lVert\Delta W\rVert_F$ grows markedly
from the early to the late checkpoint, showing that the generator
learns to apply progressively stronger updates rather than converging to a
conservative solution. Second, the updates develop a clear structure across
contexts rather than collapsing onto a single shared adapter. While the input-side
(gate and up) projections become increasingly consistent across contexts, the
output-side (down) projections move in the opposite direction, and as shown in
\Cref{fig:lora_down_hist}, the distribution of the down-projection
cross-context cosine shifts noticeably toward \emph{lower} values from the early to
the late checkpoint. This shows that CLIK gradually develops a shared, largely
context-independent backbone in its input-side adapters, while pushing the down
projections to carry progressively more context-specific information.

This consolidation is not uniform across module types, and instead reveals a clear
functional specialization (\Cref{fig:lora_per_layer,fig:lora_scatter,fig:lora_violin}). The gate and up projections
exhibit high cross-context cosine throughout the network, and this value increases
slightly with training, showing that the generator converges toward a stable,
nearly context-independent transformation for these input-side modules. In
contrast, the down projections show markedly lower cosine that is strongly
layer-dependent, and, unlike the other two kinds, their cross-context cosine
\emph{decreases} with training (mirroring the leftward shift of the down-projection
distribution in \Cref{fig:lora_down_hist}), with the largest changes
concentrated in the deep layers. The scatter plot mirrors this pattern, as gate and
up modules cluster tightly near the upper-right corner above the identity line,
whereas down modules are widely dispersed and frequently fall below it. Overall,
these observations show that CLIK learns a division of labor in which the
input-side projections form a consistent, shared backbone of adaptation, while the
output-side (down) projections, particularly in deeper layers, carry the
context-specific information needed to internalize each individual context. Because
all of these quantities are computed on $\Delta W$, they are invariant to the
reparameterization $(\mathbf{B}, \mathbf{A}) \mapsto (\mathbf{B}\mathbf{R},
\mathbf{R}^{-1}\mathbf{A})$ for any invertible $\mathbf{R} \in \mathbb{R}^{r\times r}$
and do not depend on how the update is split into factors.

\paragraph{Internalized Knowledge Structure in LoRA.}

To examine how the internalized information is distributed over the factors of the
adapter, we intervene on the generated adapter before inference and measure the
resulting change on SQuAD (\Cref{tab:knowledge}). Substituting
$\mathrm{down}_A$ with the factor generated for a different context leaves
performance essentially unchanged, whereas applying the same substitution to
$\mathrm{down}_B$ causes a near-complete collapse, and swapping the full adapter is only
marginally more damaging than swapping $\mathrm{down}_B$ alone. Substituting the gate and up
factors across contexts, by contrast, has only a minor effect, which shows that
these input-side factors are largely shared rather than context-dependent. Unlike the
analyses above, these single-factor interventions depend on the coordinate system the
generator uses. The $\mathrm{down}_A$ result is meaningful even so: in the
generator's own coordinates, $\mathrm{down}_A$ is interchangeable across contexts. The
$\mathrm{down}_B$ result, however, does not by itself show that the context is stored
in $\mathrm{down}_B$, because pairing the $\mathrm{down}_B$ of one context with the
$\mathrm{down}_A$ of another combines factors expressed in different coordinates.

\begin{wraptable}{r}{0.52\linewidth}
    \centering
    \vspace{-\baselineskip}
    \tablestyle{5pt}{1.15}
    \caption{Intervention study on SQuAD (Precision / Recall / F1). Each generated
    adapter is perturbed before inference.}
    \label{tab:knowledge}
    \begin{tabular}{@{}l ccc@{}}
        \shline
        \noalign{\vspace{1pt}}
        Intervention & Precision & Recall & F1 Score \\
        \noalign{\vspace{1pt}}
        \shline
        \noalign{\vspace{1pt}}
         Baseline & 39.4 & 81.7 & 46.3 \\
        \hline
        \noalign{\vspace{1pt}}
        \rowcolor{gray!10} \multicolumn{4}{c}{\textit{Swap across contexts}} \\
        \quad gate \& up        & 33.5 & 76.2 & 40.5 \\
        \quad $\mathrm{down}_A$  & 38.1 & 81.2 & 44.9 \\
        \quad $\mathrm{down}_B$  & 6.2 & 21.2 & 8.1 \\
        \quad full LoRA          & 5.4 & 19.7 & 7.2 \\
        \hline
        \noalign{\vspace{1pt}}
        \rowcolor{gray!10} \multicolumn{4}{c}{\textit{Corrupt $\mathrm{down}_B$ (destructive controls)}} \\
        \quad Gaussian noise     & 12.0 & 24.5 & 13.3 \\
        \quad shuffle entries    & 11.8 & 24.7 & 13.2 \\
        \hline
        \noalign{\vspace{1pt}}
        \rowcolor{gray!10} \multicolumn{4}{c}{\textit{Remove}} \\
        \quad gate \& up         & 11.5 & 29.2 & 13.2 \\
        \quad layers 0--15       & 19.0 & 39.0 & 21.1 \\
        \quad layers 20--34      & 14.0 & 61.3 & 19.7 \\
        \noalign{\vspace{1pt}}
        \shline
    \end{tabular}
    \vspace{-\baselineskip}
\end{wraptable}
Replacing $\mathrm{down}_B$ with scale-matched Gaussian noise or randomly permuting
its entries is as harmful as removing the adaptation altogether, which shows that the
specific learned values matter rather than the overall magnitude or sparsity
pattern.
Removing the gate and up factors is likewise detrimental, because although they are
not context-specific, the rest of the adapter was generated jointly with them and is
not usable without them. Finally, ablating groups of layers shows that the
internalized context is not confined to one depth band. The two groups are removed
separately, and the layers between them as well as the last layer are kept in both
cases. Removing the deeper layers is slightly more harmful in F1 than removing the
shallower layers, and the two removals fail differently: removing the shallower
layers mainly lowers recall, whereas removing the deeper layers mainly lowers
precision while keeping recall comparatively high.
Overall, these results show that the context-dependent part of the adapter is
compact and specific: the gate and up updates are largely shared across contexts,
$\mathrm{down}_A$ is interchangeable across contexts in the generator's coordinates,
and the context-dependent variation is expressed through the down projection.

\paragraph{From analysis to architecture.} Neither the gauge-invariant analyses nor
the factor interventions establish that the knowledge intrinsically resides in
$\mathrm{down}_B$. The constructive test is CLIK-B, which fixes a single learned
coordinate system shared by all contexts and generates only $\mathrm{down}_B$. Its
results in \Cref{tab:clikb,tab:controlled} show how much of the
performance of CLIK a per-context $\mathrm{down}_B$ alone can retain, at a much
smaller per-context state. CLIK-B retains most of the performance of CLIK, outperforms
all baselines of \Cref{tab:main} under the format instruction, and achieves a higher
average F1 than D2L-L and SHINE-L in \Cref{tab:controlled}, which confirms that a single shared
coordinate system with a per-context $\mathrm{down}_B$ captures most of the internalized
context. The remaining gap, which is largest on DROP and the multi-hop datasets, shows
that the gate, up, and $\mathrm{down}_A$ factors of CLIK also contribute
context-specific information.

\section{Conclusion}
\label{sec:conclusion}
We presented \textbf{CLIK}, a parameter generator that adapts a frozen LLM by mapping a
context directly to LoRA parameters in a single forward pass. Rather than keeping
task knowledge in the input or optimizing the model separately for each task, CLIK
internalizes the context into the parameters with a Transformer that attends from
the context to the adapter weights through cross-attention. Once the generated
adapter is inserted, downstream queries are answered from the query alone, without
re-reading the original context and without any test-time training. On document
question answering, CLIK improves clearly over the no-context base model and
compares favorably with existing hypernetwork-based methods under the same setting,
while remaining behind the base model that reads the context directly and avoiding the
per-task optimization of fine-tuning. Counterfactual probes confirm that the generated
adapter answers from the internalized context rather than from the prior of the
backbone. The same mechanism also carries over to
multi-turn dialogue, where CLIK improves over its single-turn baseline without
additional training.

\paragraph{Limitations and future work.}
Our study has limitations that define directions for future work. We evaluate
primarily on document question answering, so the behavior of CLIK on broader task
types, longer contexts, and larger backbones is not examined in this work. The generated
adapter also reflects whatever information the context provides, and we do not yet
model how to retain or reconcile knowledge across many successive updates. We view
context-to-parameter generation as a practical step toward adaptation that keeps
the flexibility of in-context learning while internalizing knowledge as fine-tuning
does, and as a basis for further work on efficient, incremental adaptation of LLMs.

\bibliographystyle{assets/plainnat}
\bibliography{paper_cite}
\appendix
\clearpage

\lstset{
  breaklines=true,
  breakatwhitespace=false,
  frame=lines,
  framesep=2mm,
  basicstyle=\fontsize{8}{8}\ttfamily\selectfont,
  columns=fullflexible,
  keepspaces=true,
  showstringspaces=false,
  aboveskip=0.5em,
  belowskip=0.5em,
}

\definecolor{gtclr}{HTML}{1B7F3B}     
\definecolor{baseclr}{HTML}{C0392B}   
\definecolor{bctxclr}{HTML}{0E7C7B}   
\definecolor{d2lclr}{HTML}{D35400}    
\definecolor{clikclr}{HTML}{6C3483}   
\definecolor{usrclr}{HTML}{1F4E79}    
\definecolor{modelclr}{HTML}{6C3483}  
\definecolor{metaclr}{HTML}{B9770E}   
\definecolor{histclr}{HTML}{7F8C8D}   

\section{Parameter Generator Configuration}
\label{app:pg_config}

We instantiate the parameter generator described in \Cref{sec:generator} for the
Qwen3-4B backbone. \Cref{tab:pg_config} lists the full configuration; below we
relate it to the notation used in the main text.

\paragraph{Adapted modules.} The generator adapts only the three feed-forward
projections of every transformer layer---the gate, up, and down projections of
the MLP block---while the attention projections are left untouched. With a
backbone of $L=36$ layers, this yields $3L$ adapted projections. Each adapter
has rank $r=16$ and scale $\alpha=32$ (so $\alpha/r=2$ in \Cref{eq:inject}).
Following the backbone's hidden size $d=2560$ and feed-forward intermediate size
$9728$, the gate and up projections receive factors
$\mathbf{A}\in\mathbb{R}^{r\times 2560}$, $\mathbf{B}\in\mathbb{R}^{9728\times r}$,
and the down projection receives
$\mathbf{A}\in\mathbb{R}^{r\times 9728}$, $\mathbf{B}\in\mathbb{R}^{2560\times r}$.
Each projection thus carries $r\,(2560+9728) = 196{,}608$ generated parameters, and the
full adapter carries $36 \times 3 \times 196{,}608 = 21{,}233{,}664 \approx 21.2$M
parameters per context, i.e., $40.5$\,MiB in BF16. For CLIK-B only the down-projection
$\mathbf{B}$ factors are generated, i.e., $36 \times 2560 \times 16 = 1{,}474{,}560$
parameters ($2.8$\,MiB), a $14.4\times$ reduction. Rank $16$ is used in all experiments.

\paragraph{Conditioning.} The all-layer backbone hidden states
(\Cref{sec:context-encoding}), of dimension $d=2560$, are combined by the
learned ELMo weighting and linearly projected to the generator width
$d_g=2048$, which is the dimension at which the generator operates.

\paragraph{Generator transformer.} The generator is a $12$-block Neural Network
Transformer with hidden width $d_g=2048$ and attention head dimension $128$
(i.e.\ $16$ heads), using the factorized across-layer / within-layer /
cross-attention pattern of \Cref{sec:generator}. Weight tokens have size
$r \times c = 16 \times 512$, so each layer is covered by $72$ tokens ($5$ for each
of the three factors with a dimension of $2560$ and $19$ for each of the three
factors with a dimension of $9728$). Inside the generator, every four consecutive rows of a token are
concatenated into one $d_g$-dimensional vector (channel accumulation $4$), which
yields $288$ generator tokens per layer. These tokens are initialized as the sum
of a learned within-layer and a learned layer position embedding, and after the
last block two output heads, one for $\mathbf{A}$ and one for $\mathbf{B}$ tokens,
project each row back to the $512$-dimensional weight space that is detokenized
into the factor matrices. The $\mathbf{B}$ head is initialized to zero, so that the
generated adapter is zero at the start of training, and the output is scaled by
$10^{-2}$.

\paragraph{Generator size.} The generator is large relative to the adapter it
produces. Each block contains an across-layer, a within-layer, and a
cross-attention sublayer, which together account for $604$M parameters over the
$12$ blocks, and a non-gated SiLU feed-forward sublayer of width $4d_g=8192$,
which accounts for $12 \times 2 \times 2048 \times 8192 \approx 403$M parameters.
With the input projection, the position embeddings, the output heads, and the
normalization layers, the generator has $1.01$B parameters in total
(\Cref{tab:pg_config}), which is the quantity matched against the baselines in
\Cref{sec:controlled}.

\begin{table}[ht]
\centering
\caption{Parameter generator configuration for the Qwen3-4B backbone.}
\label{tab:pg_config}
\begin{tabular}{lc}
\toprule
\textbf{Component} & \textbf{Value} \\
\midrule
\rowcolor{gray!10} \multicolumn{2}{c}{\emph{Backbone (Qwen3-4B-Instruct-2507)}} \\
Hidden size $d$ & $2560$ \\
Number of layers $L$ & $36$ \\
FFN intermediate size & $9728$ \\
\midrule
\rowcolor{gray!10} \multicolumn{2}{c}{\emph{Generated adapter (LoRA)}} \\
Rank $r$ & $16$ \\
Scale $\alpha$ & $32$ \\
Target modules & \texttt{mlp.\{gate,up,down\}\_proj} \\
Adapted layers & all $36$ \\
Generated parameters per context (CLIK) & $21.2$M ($40.5$\,MiB, BF16) \\
Generated parameters per context (CLIK-B) & $1.47$M ($2.8$\,MiB, BF16) \\
\midrule
\rowcolor{gray!10} \multicolumn{2}{c}{\emph{Generator network}} \\
Transformer blocks & $12$ \\
Hidden width $d_g$ & $2048$ \\
Attention head dim (heads) & $128$ ($16$) \\
Feed-forward width & $4\,d_g$ \\
Token block width $c$ & $512$ \\
Channel accumulation & $4$ \\
Output dimension & $512$ \\
Feed-forward type & non-gated (SiLU) \\
Parameters: attention sublayers & $604.3$M \\
Parameters: feed-forward sublayers & $402.8$M \\
Parameters: input projection, embeddings, heads, norms & $6.5$M \\
Parameters: total & $1{,}013.5$M \\
\midrule
\rowcolor{gray!10} \multicolumn{2}{c}{\emph{Training}} \\
Training compute (GPU-hours) & $3.9\times10^4$ ($25.5$ days) \\
\bottomrule
\end{tabular}
\end{table}

\section{Data Generation Configuration}
\label{app:prompts}

To construct the training corpus described in \Cref{sec:data}, we prompt
the frozen backbone to synthesize question--answer pairs grounded in each
FineWeb-Edu passage. The generation is driven by a fixed system prompt
(\Cref{code:qa_system_prompt}) that frames the task as reading-comprehension
question generation, together with a user instruction template
(\Cref{code:qa_user_prompt}) that injects the passage and specifies the
formatting rules for both questions and answers. We deliberately leave the
answer length and response style unconstrained, so that the synthesized pairs
follow the model's native in-context behavior.

\begin{listing}[ht]
\begin{lstlisting}
SYSTEM_PROMPT = (
    "You are a creative and helpful assistant.\n"
    "You are tasked with generating questions for reading comprehension tests.\n"
    "You will be given a context and you need to generate questions and corresponding answers from the given context.\n"
    "The questions should be highly specific to the information provided in the context, not general questions that suit any context.\n"
    "**DO NOT** hallucinate or make up information."
)
\end{lstlisting}
\caption{System prompt used for question--answer generation.}
\label{code:qa_system_prompt}
\end{listing}

\begin{listing}[ht]
\begin{lstlisting}
USER_PROMPT_TEMPLATE = (
    "### Instructions ###\n"
    "Generate questions and corresponding answers from the given context. The questions should be highly specific to the information provided in the context, not general questions that suit any context.\n\n"
    "### Context ###\n"
    "{context}\n\n\n"
    "### Rules ###\n"
    "Rules to follow when generating the questions:\n"
    "1. The questions must be specific to the given context and fully answerable from information present in the given context.\n"
    "2. Ask questions that are fact-seeking based on the information provided.\n"
    "3. Make sure the questions are clear and unambiguous.\n"
    "4. Phrases like 'based on the provided context', 'according to the context', 'in the context', etc., are **NOT ALLOWED** to appear in the questions.\n"
    "5. The questions should not overlap. They should be diverse, covering many aspects of the context.\n"
    "6. Do not give away too much information in the questions. For example, ask 'Who is X?' instead of 'Who is X that did Y?' when Y is clear from the context.\n"
    "7. Ignore the text formatting of the context, e.g., bold, italic, underline.\n"
    "8. Ignore typos, spacing, and grammatical errors in the context.\n\n"
    "Rules to follow when generating the answers:\n"
    "1. The answers must use the (implied) information provided in the context.\n"
    "2. Phrases like 'based on the provided context', 'according to the context', 'in the context', etc., are **NOT ALLOWED** to appear in the answers.\n"
    "3. Do not just copy words from the context. Answer the question in your own words.\n"
    "4. The answers should be detailed and comprehensive. Please include additional specific details from the context.\n\n"
    "Respond with {n_qa_pairs} question-answer pairs.\n"
    "Always use proper grammar and punctuation.\n"
    "Try to use different question forms and styles.\n"
    "Use simple words and make sure that the answers are clear and comprehensive.\n\n"
    "The question-answer pairs should be in the following format:\n"
    "Question 1: {{question_1}}\n"
    "Answer 1: {{answer_1}}\n"
    "Question 2: {{question_2}}\n"
    "Answer 2: {{answer_2}}\n"
    "..."
)
\end{lstlisting}
\caption{User instruction template used for question--answer generation.}
\label{code:qa_user_prompt}
\end{listing}

Applying these prompts to a FineWeb-Edu passage yields a context paired with
several self-generated question--answer pairs. \Cref{code:data_example} shows one
representative training instance; at training time each question--answer pair
forms a separate supervision target conditioned on the same context.

\begin{listing}[ht]
\begin{lstlisting}
Context:
The Hearst ranch, San Jose de Babicora, originally founded in the 18th century by the Jesuits, was located in northern Mexico and was in the pathway of the revolutionary movements of the 1910s-1920s. In 1907, at the age of twenty-four, Jim Barker went to Mexico, where he served as ranch manager of the million-acre estate in Chihuahua of Mrs. Phoebe Hearst. Described as a "smallish, wiry Irishman with a sandy complexion and sandy hair and sandy eyes," he served as a civilian scout for General John J. Pershing during the punitive expedition's search for Pancho Villa.
This interview with Mr. Barker, who died at age eighty-three in San Antonio, provides extensive primary source information about the Mexican Revolution, his involvement in it, the murder of Peter Keene, and the history of the Babicora ranch. Pancho Villa put a $10,000 price tag on the head of Mr. Barker in his attempt to capture Barker. This led to Barker's nine-day trip to escape being captured. Barker later served in World War I as head of an extensive intelligence operation in France and Germany and was a business consultant for the Hearst Enterprises until he retired in 1963.

Q1: Where was the Hearst ranch, San Jose de Babicora, originally located?
A1: The Hearst ranch, San Jose de Babicora, was originally located in northern Mexico.

Q2: What was Jim Barker's age when he first went to Mexico to work for Mrs. Phoebe Hearst?
A2: Jim Barker was 24 years old when he first went to Mexico to work for Mrs. Phoebe Hearst.

Q3: How is Jim Barker physically described in the text?
A3: Jim Barker is described as a "smallish, wiry Irishman with a sandy complexion and sandy hair and sandy eyes."

Q4: What was the reward offered for Jim Barker's capture, and who offered it?
A4: The reward offered for Jim Barker's capture was $10,000, and it was offered by Pancho Villa.

Q5: What role did Jim Barker hold during World War I?
A5: During World War I, Jim Barker served as the head of an extensive intelligence operation in France and Germany.
\end{lstlisting}
\caption{A representative FineWeb-Edu training instance: a context passage paired with self-generated question--answer pairs.}
\label{code:data_example}
\end{listing}

\section{Evaluation Configuration}
\label{app:eval_prompt}

At evaluation time, the backbone never sees the source passage: the knowledge
required to answer is supplied entirely by the LoRA adapter that the generator
produces from that passage. Under the protocol with the format instruction, each test
question is wrapped in the fixed answer-format instruction shown in
\Cref{code:eval_prompt}, which is the instruction that Doc-to-LoRA~\citep{charakorn2026doc}
uses for SQuAD, DROP, and ROPES; under the protocol without the format instruction, the
question is given alone. In both cases the prompt is wrapped in the backbone's chat
template, passed to the adapted model, and decoded greedily (\Cref{eq:infer})
with at most $256$ new tokens. The answer-format instruction is used only at
evaluation time: training prompts contain the question alone, wrapped in the same
chat template (\Cref{sec:data}), so the protocol without the format instruction matches
the training format, whereas the protocol with it is out of distribution for the
generator and tests whether the adapted model still follows an instruction it was not
trained with. We deliberately provide
neither the document nor any in-context exemplar, so that the context can only reach
the model through the generated parameters.
\begin{listing}[ht]
\begin{lstlisting}
EVAL_PROMPT_TEMPLATE = (
    "Answer the following question. Output only the answer and do not output any other words.\n\n"
    "Question: {input}"
)
\end{lstlisting}
\caption{Instruction template used to query the adapted model at evaluation time.}
\label{code:eval_prompt}
\end{listing}

\section{Re-implemented Baselines}
\label{app:baselines}

We re-train Doc-to-LoRA and SHINE from their official implementations. In
\Cref{tab:main}, all generators share the backbone (Qwen3-4B-Instruct, frozen), the
training data (the same FineWeb-Edu passages and synthesized question--answer pairs),
and the generated adapter (rank $16$, $\alpha=32$, gate, up, and down projections of all
$36$ layers), while each baseline keeps its own architecture and training objective:
Doc-to-LoRA is trained by context distillation with a KL objective and an $\ell_1$
penalty on the generated adapter, and SHINE by next-token prediction on the answers.
Because SHINE maps each memory state of dimension $d=2560$ directly to a slice of the
adapter, its $589{,}824$ generated parameters per layer require $\lceil 589{,}824/2560
\rceil = 231$ memory states per layer, and the $1{,}536$ surplus outputs are discarded.
All other architectural and training hyperparameters follow the official
implementations. The controlled comparison of \Cref{sec:controlled} varies only the
generator scale and keeps the objective and all other settings of each method as in
\Cref{tab:main}, so the native-size models are D2L and SHINE in \Cref{tab:controlled}.
\Cref{tab:baseline_config} lists the architecture-specific settings. The parameter
count of Doc-to-LoRA includes its per-module output head and the context-independent
LoRA biases, and that of SHINE includes the rank-$128$ meta-LoRA ($0.26$B), the memory
tokens, and the M2P Transformer ($0.21$B). D2L-L widens the latent dimension
of Doc-to-LoRA to $576$, and SHINE-L deepens the M2P Transformer to $14$ layers.

\begin{table}[ht]
\centering
\caption{Architecture-specific settings of the re-implemented baselines.}
\label{tab:baseline_config}
\resizebox{\linewidth}{!}{%
\begin{tabular}{lccc}
\toprule
 & Doc-to-LoRA & SHINE & CLIK \\
\midrule
Context encoder & frozen backbone & backbone + trainable LoRA (rank $128$) & frozen backbone \\
Context representation & per-layer activations & $231$ memory states per layer & ELMo mixture, full sequence \\
Generator & Perceiver per layer & alternating across-layer / & factorized attention \\
 & ($8$ latents, $9$ blocks, width $512$) & within-layer Transformer & + cross-attention \\
Generator blocks / width & $9$ / $512$ & $4$ / $2560$ & $12$ / $2048$ \\
Interaction across layers & none & across memory states & across weight tokens \\
Objective & KL distillation + $\ell_1$ & next-token prediction & KL distillation \\
Generator parameters (native) & $0.89$B & $0.48$B & $1.01$B \\
Generator parameters (scaled) & $1.01$B & $1.00$B & n/a \\
Training GPU-hours (native / scaled) & $13.0$k / $13.1$k & $13.6$k / $23.7$k & $39.1$k \\
\bottomrule
\end{tabular}%
}
\end{table}

\section{Training and Inference Cost}
\label{app:cost}

Training costs in \Cref{sec:controlled} are reported in GPU-hours. The backbone has $4.02$B parameters ($8.0$\,GB in
BF16), and its KV cache takes $144$\,KiB per token. To control the variables across
methods and make the different strategies directly comparable, every query and every
answer is fixed to $32$ tokens in all inference measurements, including the multi-turn
latencies.

\paragraph{Training.} The training set consists of the $9.67$M passages of
FineWeb-Edu, each truncated to $2048$ tokens, with $5$ question--answer pairs per
passage and $2$ epochs, i.e., $96.7$M training examples. Training CLIK takes
$3.9\times10^4$ GPU-hours, i.e., $25.5$ days. Doc-to-LoRA and SHINE take
$1.30\times10^4$ and $1.36\times10^4$ GPU-hours ($8.5$ and $8.9$ days), and CLIK-B,
whose generator processes only $20$ tokens per layer, takes $1.03\times10^4$ GPU-hours
($6.7$ days).

\paragraph{Inference.} Adapters that are applied per request use segmented
multi-adapter kernels. Internalization of the context-based
methods is one prefill of the context, followed for SnapKV by scoring with the
observation window. That scoring is counted once in the internalization column of \Cref{tab:efficiency}. The SnapKV
accuracy reported there uses a cache selected for the current question, so a new
question selects again. The \emph{New question} column is this added latency: the TTFT
when the internalized state is reused, and the internalization time plus the TTFT for
SnapKV. LLMLingua-2 runs an XLM-RoBERTa-large token classifier over the
context, including tokenization and word aggregation on the CPU, and then passes the
compressed $512$-token prompt to the backbone for every query. Per-context states are
loaded from host memory. Peak memory includes the backbone and the generator or
compressor weights, the $32$ per-context states and the KV cache of the batched run, the
workspace, and the transient activations of one internalization. For the multi-turn latencies in \Cref{tab:multiturn_cost}, Dynamic LoRA
re-internalizes the initial context together with the four previous turns before
answering.

\section{Construction of the Counterfactual Probes}
\label{app:probes}

\paragraph{Paired counterfactual documents.} We sample passages from the evaluation
sets and, for each, select one answer-bearing fact (an entity, number, or date). We
create two copies that differ only in this fact, each replaced by a distinct nonce
value of the same type (e.g., invented names or dates that do not occur in the
original passage), and keep the question unchanged. Each copy is internalized
separately, and both adapters are queried with the identical question.

\paragraph{Novel facts.} We generate short passages about fictional entities (invented
people, organizations, and places with nonce names) together with fact-seeking
questions, and verify that Base answers them at chance level.

\paragraph{Strong prior conflict.} We edit passages so that they contradict a fact
that Base answers correctly without context (e.g., the location of a well-known
landmark), and ask for that fact. An answer is counted as \emph{Ctx} if it matches the
edited value and as \emph{Prior} if it matches the original value.

\paragraph{Wrong-context swap and unrelated queries.} For the wrong-context swap, each
question is asked under the adapter of a randomly chosen passage from a different
evaluation example. For unrelated-query interference, we take general-knowledge
questions that Base answers correctly without any adapter, ask them under the adapter
of a random passage, and report accuracy relative to Base.

\section{Frequently Asked Questions}
\label{app:faq}

\paragraph{What distinguishes CLIK from SHINE?}
The across-layer and within-layer attention of CLIK is the factorized attention of the
Neural Network Transformer~\citep{tencent2026hy}. SHINE already attends in both
directions: \Cref{tab:baseline_config} lists its M2P Transformer as alternating
across-layer and within-layer attention over a fixed set of memory states. The
difference is the representation those layers read. SHINE first trains a LoRA that
compresses the context into a fixed-length token sequence, and that sequence is the
initial input of the parameter generator. CLIK leaves the context at its full length
and supplies it as the input of cross-attention, so each weight token reads the
context while the adapter is generated. This is the change that raises accuracy over
SHINE. \Cref{tab:arch_ablation} shows that the two attentions improve CLIK, and the
margin over SHINE is the margin between cross-attention over the full context
representation and generation from a fixed-length memory. Per-layer generation with
no interaction across layers is the design of Doc-to-LoRA, as recorded in the same
table.

\paragraph{Why is CLIK-B absent from the counterfactual probes?}
In CLIK-B the gate and up factors are shared across contexts and mainly carry the
pattern of the question-answering task, while the per-context
$\mathbf{B}^{\mathrm{down}}$ mainly compresses the context. On instruction
following, and on tests that compress a context outside question answering, this
split lowers accuracy substantially, so CLIK-B remains a compact variant for the
question-answering setting and is omitted from \Cref{tab:counterfactual}. The
published Doc-to-LoRA makes the same split, generating only the down-projection
LoRA, and reproducing that version follows the supplied context poorly. The
Doc-to-LoRA compared in this paper generates the gate, up, and down factors of
every layer, the same adapter CLIK generates, and that full adapter follows the
context.

\paragraph{Does the small in-domain gain rule out overlap with the benchmarks?}
Context following and corpus overlap are separate questions. The probes in
\Cref{tab:counterfactual} test whether the answer comes from the given context. Adding
the training splits of SQuAD, DROP, and ROPES (\Cref{tab:indomain}) measures how much
in-domain supervision changes F1, and a small change there does not test overlap.
The training passages are the FineWeb data provided with Doc-to-LoRA, and the
question--answer pairs are synthesized again for this paper, so those pairs do not
overlap the evaluation questions and answers. The training mixture in the
Doc-to-LoRA paper is that cleaned FineWeb together with the training splits of
SQuAD, ROPES, and DROP, so the passages do not overlap the evaluation contexts
either.

\paragraph{SnapKV decodes faster than CLIK. Where does a new question change the cost?}
The throughput in \Cref{tab:efficiency} already includes attention over the retained
state: SnapKV decodes at $9.0$k tokens per second and CLIK at $8.3$k. SnapKV keeps
tokens according to an observation window at the end of the prompt, and on these
datasets that window is the question, so the reported F1 uses a cache selected for
the current question. The internalization column counts that selection once. A
further question on the same document selects the cache again, while CLIK generates
the adapter once and reuses it for every question. The \emph{New question} column is
this added latency. For SnapKV it is the internalization time plus the time to first
token; for CLIK it is the time to first token alone.

\paragraph{The data prompt asks for detailed answers. Is the short-answer comparison fair?}
The generator is trained to reproduce in-context behavior without being required to
emit a short answer, so the supervision is the long question--answer pairs of
\Cref{code:qa_user_prompt}. That prompt asks for a detailed and comprehensive answer
and for additional details from the context, which is how the long answers are
obtained. When an evaluation requires a short answer, the format instruction of
\Cref{code:eval_prompt} is added at test time. Every baseline is trained on the same
pairs and scored under the same protocol, so the answer length is shared by the
comparison.

\paragraph{How much of the gain is the fact, and how much is the answer format?}
\Cref{tab:split} separates the two by precision and recall. Under the short-answer
instruction, precision mainly measures whether the output follows that instruction.
Training uses the same long-answer instruction on every example and never includes
the short-answer instruction, so on this precision CLIK remains behind Base + Context.
Without the short-answer instruction, the precision and recall of CLIK stay close to
those of Base + Context.

\section{Qualitative Comparison}
\label{app:qualitative}

To illustrate the behavioral differences between methods, we present a
side-by-side comparison on a single SQuAD passage. For every question we report
four answers: \textbf{Base} (the frozen backbone queried with the question
only, i.e., closed-book and without the passage), \textbf{Base+Context} (the
backbone given the passage in-context, an upper bound that has direct access to
the evidence), \textbf{Doc-to-LoRA} (the document-to-LoRA baseline), and
\textbf{CLIK} (ours). Both Doc-to-LoRA and CLIK answer from the LoRA-internalized
knowledge without seeing the passage at inference time. \Cref{code:qual_example}
shows the resulting transcript. The closed-book Base model already answers several
questions correctly from its prior, since Super Bowl~50 is well known, but adds
spurious or incorrect details (e.g., in Q2). CLIK produces short answers that agree
with the in-context reference on Q1--Q4 but answers Q5 incorrectly (``Super Bowl 50''
instead of ``Super Bowl''). Because the backbone already knows much of this passage,
this example illustrates output behavior rather than knowledge acquisition, which is
tested with counterfactual probes in \Cref{sec:counterfactual}.

\begin{listing}[ht]
\begin{Verbatim}[breaklines, breakanywhere, frame=lines, framesep=2mm, commandchars=\\\{\}, fontsize={\fontsize{8}{8}\selectfont}]
Context:
Super Bowl 50 was an American football game to determine the champion of the National Football League (NFL) for the 2015 season. The American Football Conference (AFC) champion Denver Broncos defeated the National Football Conference (NFC) champion Carolina Panthers 24-10 to earn their third Super Bowl title. The game was played on February 7, 2016, at Levi's Stadium in the San Francisco Bay Area at Santa Clara, California. As this was the 50th Super Bowl, the league emphasized the "golden anniversary" with various gold-themed initiatives, as well as temporarily suspending the tradition of naming each Super Bowl game with Roman numerals (under which the game would have been known as "Super Bowl L"), so that the logo could prominently feature the Arabic numerals 50.

====================================================================
Q1: Where did Super Bowl 50 take place?
\textcolor{gtclr}{[Ground Truth]  Santa Clara, California}
\textcolor{baseclr}{[Base]          Super Bowl 50 took place at Levi's Stadium in Santa Clara, California.}
\textcolor{bctxclr}{[Base+Context]  Super Bowl 50 took place at Levi's Stadium in Santa Clara, California.}
\textcolor{d2lclr}{[Doc-to-LoRA]   Super Bowl 50 took place at Levi's Stadium in Santa Clara, California, in the San Francisco Bay Area.}
\textcolor{clikclr}{[CLIK]          Super Bowl 50 took place at Levi's Stadium in the San Francisco Bay Area at Santa Clara, California.}

====================================================================
Q2: What color was used to emphasize the 50th anniversary of the Super Bowl?
\textcolor{gtclr}{[Ground Truth]  gold}
\textcolor{baseclr}{[Base]          The color used to emphasize the 50th anniversary of the Super Bowl was gold. The 50th Super Bowl (Super Bowl XLV) was held in 2011, and while the official anniversary of the first Super Bowl (1967) was celebrated in 2017, the 50th Super Bowl (Super Bowl LVI) in 2022 marked the 50th anniversary of the first Super Bowl. For that celebration, gold was prominently used in branding, logos, and promotional materials to highlight the milestone.}
\textcolor{bctxclr}{[Base+Context]  The color used to emphasize the 50th anniversary of the Super Bowl was gold. This is evident from the context, which states: "the league emphasized the 'golden anniversary' with various gold-themed initiatives."}
\textcolor{d2lclr}{[Doc-to-LoRA]   Gold was used to emphasize the 50th anniversary of the Super Bowl.}
\textcolor{clikclr}{[CLIK]          The color was gold.}

====================================================================
Q3: If Roman numerals were used, what would Super Bowl 50 have been called?
\textcolor{gtclr}{[Ground Truth]  Super Bowl L}
\textcolor{baseclr}{[Base]          To find out what Super Bowl 50 would have been called in Roman numerals: the number 50 is represented as L, so Super Bowl 50 in Roman numerals would be called Super Bowl L.}
\textcolor{bctxclr}{[Base+Context]  If Roman numerals were used, Super Bowl 50 would have been called Super Bowl L. In Roman numerals, 50 is represented by L. However, for the 50th Super Bowl the NFL temporarily suspended the use of Roman numerals and instead used the Arabic numeral 50.}
\textcolor{d2lclr}{[Doc-to-LoRA]   If Roman numerals were used, Super Bowl 50 would have been called "Super Bowl L".}
\textcolor{clikclr}{[CLIK]          Super Bowl L}

====================================================================
Q4: Where was Super Bowl 50 held?
\textcolor{gtclr}{[Ground Truth]  Santa Clara, California}
\textcolor{baseclr}{[Base]          Super Bowl 50 was held at Levi's Stadium in Santa Clara, California.}
\textcolor{bctxclr}{[Base+Context]  Super Bowl 50 was held at Levi's Stadium in Santa Clara, California.}
\textcolor{d2lclr}{[Doc-to-LoRA]   Super Bowl 50 was held at Levi's Stadium in Santa Clara, California, in the San Francisco Bay Area.}
\textcolor{clikclr}{[CLIK]          Super Bowl 50 was held at Levi's Stadium in the San Francisco Bay Area at Santa Clara, California.}

====================================================================
Q5: What is the name of the NFL championship game?
\textcolor{gtclr}{[Ground Truth]  Super Bowl}
\textcolor{baseclr}{[Base]          The name of the NFL championship game is the Super Bowl, the annual championship game of the National Football League where the winners of the AFC and NFC face off to determine the league champion.}
\textcolor{bctxclr}{[Base+Context]  The name of the NFL championship game is the Super Bowl.}
\textcolor{d2lclr}{[Doc-to-LoRA]   The name of the NFL championship game is Super Bowl LVI.}
\textcolor{clikclr}{[CLIK]          Super Bowl 50}
\end{Verbatim}
\caption{Qualitative comparison on a SQuAD passage about Super Bowl~50. Answers are color-coded by source: \textcolor{gtclr}{Ground Truth}, \textcolor{baseclr}{Base} (closed-book, no passage), \textcolor{bctxclr}{Base+Context} (passage given in-context, upper bound), \textcolor{d2lclr}{Doc-to-LoRA}, and \textcolor{clikclr}{CLIK} (ours). Doc-to-LoRA and CLIK answer from the internalized LoRA adapter without the passage. CLIK agrees with the in-context reference on Q1--Q4 and errs on Q5, while the closed-book Base often adds unsupported or incorrect details.}
\label{code:qual_example}
\end{listing}

\section{Multi-turn Dialogue Example}
\label{app:multiturn}

We illustrate the two multi-turn mechanisms introduced in
\Cref{sec:multiturn} on a shared initial context $c_0$ (a SQuAD passage
about Super Bowl~50) and the same sequence of five questions. To keep the
listings compact, the passage is shown once and abbreviated as $c_0$ thereafter;
each turn reports the base model's input and its output. The two scenarios differ
in \emph{what grows} as the dialogue proceeds. In the \textbf{dynamic KV cache} scenario (\Cref{code:mt_kv}), the adapter is
generated once from $c_0$ and \emph{kept fixed}; conversational state is carried
by the base model's KV cache, so the model input \emph{accumulates} the previous
question--answer pairs turn by turn. In the \textbf{dynamic LoRA} scenario
(\Cref{code:mt_lora}), the KV cache is discarded and the model input always
contains only the current question; instead, the context fed to the parameter
generator \emph{grows} ($c_0 \!\to\! c_1 \!\to\! \dots$) and a fresh adapter is
synthesized at every turn.

\begin{listing}[ht]
\begin{Verbatim}[breaklines, breakanywhere, frame=lines, framesep=2mm, commandchars=\\\{\}, fontsize={\fontsize{8}{8}\selectfont}]
\textcolor{metaclr}{Adapter (fixed): LoRA = PG(c0)   -- internalized once; passage never enters the input}
\textcolor{metaclr}{c0 = "Super Bowl 50 was an American football game ... so that the logo could prominently feature the Arabic numerals 50."}

\textbf{===== Turn 1 =====}
Model input : \textcolor{usrclr}{Q1: Which NFL team represented the AFC at Super Bowl 50?}
Model output: \textcolor{modelclr}{Denver Broncos}

\textbf{===== Turn 2 =====}  (KV cache from Turn 1 retained)
Model input : \textcolor{histclr}{Q1: Which NFL team represented the AFC at Super Bowl 50?}
              \textcolor{histclr}{A1: Denver Broncos}
              \textcolor{usrclr}{Q2: Which NFL team represented the NFC at Super Bowl 50?}
Model output: \textcolor{modelclr}{Carolina Panthers}

\textbf{===== Turn 3 =====}  (KV cache from Turns 1-2 retained)
Model input : \textcolor{histclr}{Q1: Which NFL team represented the AFC at Super Bowl 50?}
              \textcolor{histclr}{A1: Denver Broncos}
              \textcolor{histclr}{Q2: Which NFL team represented the NFC at Super Bowl 50?}
              \textcolor{histclr}{A2: Carolina Panthers}
              \textcolor{usrclr}{Q3: Where did Super Bowl 50 take place?}
Model output: \textcolor{modelclr}{Levi's Stadium in the San Francisco Bay Area at Santa Clara, California}

\textbf{===== Turn 4 =====}  (KV cache from Turns 1-3 retained)
Model input : \textcolor{histclr}{... accumulated (Q1,A1) (Q2,A2) (Q3,A3) ...}
              \textcolor{usrclr}{Q4: Which NFL team won Super Bowl 50?}
Model output: \textcolor{modelclr}{Denver Broncos}

\textbf{===== Turn 5 =====}  (KV cache from Turns 1-4 retained)
Model input : \textcolor{histclr}{... accumulated (Q1,A1) ... (Q4,A4) ...}
              \textcolor{usrclr}{Q5: What color was used to emphasize the 50th anniversary of the Super Bowl?}
Model output: \textcolor{modelclr}{Gold}
\end{Verbatim}
\caption{\textbf{Dynamic KV cache.} The adapter \textcolor{metaclr}{LoRA = PG($c_0$)} is generated once and held fixed. Conversational state is propagated through the base model's KV cache, so the \textcolor{usrclr}{model input} \emph{accumulates} the previous \textcolor{histclr}{question--answer history} turn by turn, while the \textcolor{modelclr}{model output} is produced from the fixed adapter plus this growing history.}
\label{code:mt_kv}
\end{listing}

\begin{listing}[ht]
\begin{Verbatim}[breaklines, breakanywhere, frame=lines, framesep=2mm, commandchars=\\\{\}, fontsize={\fontsize{8}{8}\selectfont}]
\textcolor{metaclr}{c0 = "Super Bowl 50 was an American football game ... so that the logo could prominently feature the Arabic numerals 50."}

\textbf{===== Turn 1 =====}
Adapter     : \textcolor{metaclr}{LoRA = PG(c0)}
Model input : \textcolor{usrclr}{Q1: Which NFL team represented the AFC at Super Bowl 50?}
Model output: \textcolor{modelclr}{Denver Broncos}

\textbf{===== Turn 2 =====}
Adapter     : \textcolor{metaclr}{LoRA = PG(c1),  c1 = c0 + (Q1, A1)}
Model input : \textcolor{usrclr}{Q2: Which NFL team represented the NFC at Super Bowl 50?}
Model output: \textcolor{modelclr}{Carolina Panthers}

\textbf{===== Turn 3 =====}
Adapter     : \textcolor{metaclr}{LoRA = PG(c2),  c2 = c1 + (Q2, A2)}
Model input : \textcolor{usrclr}{Q3: Where did Super Bowl 50 take place?}
Model output: \textcolor{modelclr}{Santa Clara, California}

\textbf{===== Turn 4 =====}
Adapter     : \textcolor{metaclr}{LoRA = PG(c3),  c3 = c2 + (Q3, A3)}
Model input : \textcolor{usrclr}{Q4: Which NFL team won Super Bowl 50?}
Model output: \textcolor{modelclr}{Denver Broncos}

\textbf{===== Turn 5 =====}
Adapter     : \textcolor{metaclr}{LoRA = PG(c4),  c4 = c3 + (Q4, A4)}
Model input : \textcolor{usrclr}{Q5: What color was used to emphasize the 50th anniversary of the Super Bowl?}
Model output: \textcolor{modelclr}{The color used to emphasize the 50th anniversary of the Super Bowl was gold.}
\end{Verbatim}
\caption{\textbf{Dynamic LoRA.} The KV cache is discarded, so the \textcolor{usrclr}{model input} always contains only the current question and never accumulates. Instead, the \textcolor{metaclr}{context $c_i$ fed to the parameter generator grows} ($c_0\!\to\!c_1\!\to\!\dots$, each appending the latest $(\,q,a\,)$ pair), and a fresh adapter is regenerated every turn to produce the \textcolor{modelclr}{model output}.}
\label{code:mt_lora}
\end{listing}

\end{document}